\documentclass[review, 3p, 11pt, longtitle, letter]{elsarticle} 
\usepackage{url}
\usepackage{amssymb}
\usepackage{amsmath}
\usepackage{graphicx}
\usepackage{algorithm}
\usepackage[noend]{algpseudocode}
\usepackage{subfig}
\usepackage{slashbox}

\usepackage{url}
\biboptions{authoryear}

\journal{Medical Image Analysis}

\begin{document}
\begin{frontmatter}
\title{Flow Network Tracking for Spatiotemporal \\ and Periodic Point Matching: \\ Applied to Cardiac Motion Analysis}

\author[mymainaddress]{Nripesh Parajuli\corref{mycorrespondingauthor}}
\cortext[mycorrespondingauthor]{Corresponding author}
\ead{nripesh.parajuli@yale.edu}

\author[mysecondaddress]{Allen Lu} 
\author[mysecondaddress]{Kevinminh Ta} 
\author[mythirdaddress]{John C. Stendahl}
\author[mythirdaddress]{Nabil Boutagy}
\author[mythirdaddress]{Imran Alkhalil}
\author[mythirdaddress]{Melissa Eberle}
\author[myfourthaddress]{Geng-Shi Jeng}
\author[myfifthaddress]{Maria Zontak}
\author[myfourthaddress]{Matthew O'Donnell}
\author[mythirdaddress,mysixthaddress]{Albert J. Sinusas}
\author[mymainaddress,mysecondaddress,mysixthaddress]{James S. Duncan}

\address[mymainaddress]{Department of Electrical Engineering, Yale University, New Haven, CT 06520, USA}
\address[mysecondaddress]{Department of Biomedical Engineering, Yale University, New Haven, CT 06520, USA}
\address[mythirdaddress]{Department of Internal Medicine, Yale University, New Haven, CT 06520, USA}
\address[mysixthaddress]{Department of Radiology \& Biomedical Imaging, Yale University, New Haven, CT 06520, USA}
\address[myfourthaddress]{Department of Bioengineering, Washington University, Seattle 98195, WA, USA}
\address[myfifthaddress]{College of Computer and Information Science, Northeastern University, Seattle 98195, WA, USA}

\begin{abstract}
The accurate quantification of left ventricular (LV) deformation/strain shows significant promise for quantitatively assessing cardiac function for use in diagnosis and therapy planning \citep{Jasaityte2012}. However, accurate estimation of the displacement of myocardial tissue and hence LV strain has been challenging due to a variety of issues, including those related to deriving tracking tokens from images and following tissue locations over the entire cardiac cycle. In this work, we propose a point matching scheme where correspondences are modeled as flow through a graphical network. Myocardial surface points are set up as nodes in the network and edges define neighborhood relationships temporally. The novelty lies in the constraints that are imposed on the matching scheme, which render the correspondences one-to-one through the entire cardiac cycle, and not just two consecutive frames.
The constraints also encourage motion to be cyclic, which is an important characteristic of LV motion. We validate our method by applying it to the estimation of quantitative LV displacement and strain estimation using 8 synthetic and 8 open-chested canine 4D echocardiographic image sequences, the latter with sonomicrometric crystals implanted on the LV wall. We were able to achieve excellent tracking accuracy on the synthetic dataset and observed a good correlation with crystal-based strains on the in-vivo data.

\end{abstract}


\end{frontmatter}

\section{Introduction}
\par Cardiovascular diseases (CVDs) were the leading cause of death in the world in 2013 according to the AHA - as stated in the heart disease and stroke statistics, 2017 update \citep{benjamin2017heart}. 
Among CVDs, ischemic heart diseases are the most common, which occur as a result of the formation of atherosclerotic plaques in the coronary arteries. This results in the narrowing of the arteries that can lead to an inadequate supply of blood to the left ventricle (myocardial ischemia). A sudden blockage of the arteries, for instance, due to plaque ruptures, can lead to irreversible tissue damage (myocardial infarction) and ultimately heart failure, which can be fatal.
\par Analysis of the left ventricle (LV), which is the main pumping chamber of the heart, can provide invaluable insights into cardiac health. An ischemic event in the ventricle is manifest as a reduction of the contractility of the LV wall muscle (myocardium). However, these wall motion abnormalities can be localized to a specific area of the LV and therefore, difficult to detect. Global measures of left ventricular function, such as ejection fraction (EF), are often not sensitive enough to detect these changes and do not provide important information on the location of the dysfunctional tissue. The 2013 ACC/AHA guideline on heart failure states that approximately $50\% $ of heart failure cases present themselves with preserved ejection fraction \citep{yancy20132013}. A significant proportion of these cases are ischemic diseases with wall motion abnormalities. Therefore, it is crucial that a more local and informative measure be developed and adopted. 
\par Visual wall motion scoring is a popular clinical technique for assessing such local deformation behavior and has been shown to be more predictive of clinical outcomes than EF \citep{galasko2001prospective, eek2010strain}. Stress imaging based wall motion scoring is also of interest and is shown to be of great utility in identifying and stratifying risk factors associated with mortality \citep{yao2003practical}. However, visual wall motion scoring is prone to a high level of uncertainty because it is a semi-quantitative, subjective metric, and hence, the interobserver variability has been shown to be high in a study involving multiple imaging modalities \citep{hoffmann2006analysis}. In this context, Lagrangian strain analysis has emerged as a viable method for wall motion quantification and can assist detection and diagnosis of disease, as well as track the therapy, recovery, and management process \citep{pellerin2003tissue, gotte2006myocardial}.
\par Among the popular imaging modalities that could be useful for assessing myocardial strain such as magnetic resonance imaging (MRI, e.g., \citet{le2017sparse, papademetris2002estimation}), computed tomography (CT, e.g., \citet{cury2008comprehensive, sugeng2006quantitative}) and echocardiography (echo, e.g., \citet{compas2014radial, heyde2013elastic}), echo is of special interest because of its affordability, higher frame rates compared to MR and CT, and portability. 
However, the trade-off is that ultrasound imaging is prone to artifacts such as bone shadows, attenuation, signal dropouts, incomplete geometry due to wrong imaging angle and location. These issues call for robust image analysis algorithms and processing pipeline. 
\par In this work we present a novel method aimed at improving upon prior efforts to quantify LV motion \citep{ledesma2005spatio, de2012temporal, compas2014radial} by capturing the myocardial dynamics with a fully spatiotemporal model. A spatiotemporal viewpoint is consistent with the manner in which clinical readings of echo is performed - as a movie, rather than viewing still frames. Most cardiac motion analysis methods perform frame-to-frame displacement estimation and combine the series of deformations to obtain Lagrangian displacements. Uncertainties and ambiguities, that arise at each step in time, get compounded and propagated as frame-to-frame estimations are aggregated. Therefore, significant drift can occur while tracking voxels through the cardiac cycle, particularly past the systolic phase. Another aspect of cardiac motion that is ignored by most methods is the periodicity of the deformation estimates over the cardiac cycle.
%
\par Therefore, in this work, we propose a method where the motion model accounts for global spatiotemporal consistency and correspondence as well as periodicity. We build a graphical network where myocardial surface points are set up as nodes and each node is connected to a few other nodes that are its candidate matches in the next time frame, via edges. The edges are associated with weights that capture the likelihood of a particular match. The flow $f$ through the network - a binary variable that captures whether or not a particular match amongst the candidates was chosen - is then solved via optimization and subject to a variety of constraints. 
\par We previously reported a preliminary version of flow network tracking (FNT) in \citet{parajuli2017flow}. In that work, we introduced the graphical network model for motion analysis. Here, we expand that model further to get binary-valued flow solutions in order to obtain non-overlapping and complete motion trajectories through the entire cardiac cycle. Instead of defining edge relationships by a nearest neighbor search using spatial distance, we do this by feature distance. Furthermore, by introducing additional constraints in the optimization, we are now able to encourage trajectories to be closed-looped and thereby model the periodic aspect of cardiac motion. Also in our previous effort, we used a supervised learning based Siamese network for feature learning. While that performed well with synthetic data, it was not easy to use a similar strategy with \textit{in vivo} data due to a lack of training samples. Therefore, in this work, we use an unsupervised method involving convolutional autoencoders to derive features.
\par We validate the application of our FNT shape tracking method on 8 synthetic 3D+t ultrasound image sequences developed by \citep{alessandrini2015pipeline} and on 8 open-chested canines imaged at baseline, after coronary occlusion, and with dobutamine stress (24 studies in total). Validation is done by comparing with strains obtained from implanted sonomicrometric crystals in the LV. Sonomicrometry derived strains were available for 7 baseline canine studies and 5 canine studies during ischemia and dobutamine stress. We perform a correlation analysis to compare echocardiography based strains and crystal based strains.  
\section{Related Work}
%
\subsection{Speckle/Image-based Tracking Methods}
%
%
\subsubsection{Non-rigid Registration}
Non-rigid registration methods typically consist of a model where the motion (displacement) field is parametrized by smooth functions such as B-splines. 
\citet{ledesma2005spatio} applied it to 3D ultrasound sequences in a frame-to-frame manner. \citet{heyde2012three} proposed a 3D deformation model where the LV image is transformed from Cartesian coordinate to an anatomical LV shaped coordinate system. However, most methods of this class do not use a fully spatiotemporal motion model. The optimizations are highly non-convex and are prone to get stuck in local minimums that can yield non-optimal solutions.
\par Some work has been done towards addressing the spatiotemporal alignment issue. \citet{ledesma2005spatio} proposed a 3d+t B-spline spatiotemporal model which parameterized the Lagrangian motion of a point at end diastole (ED) through the cardiac cycle . However, their model does not explicitly capture any notion of global spatiotemporal correspondence and consistency. This is because their cost function accounts for the dissimilarity with the ED frame but not with any other frames, including the neighboring frames. Also, this method does not capture large deformations effectively. \citet{de2012temporal} proposed a 3D+t diffeomorphic map-based registration method, where a B-spline parameterization over the velocity field is used. While this explicitly models the notion of spatiotemporal correspondence, the concern with velocity-based parametrization, in general, is that it is prone to error accumulation as Lagrangian displacements are calculated by integrating the velocities. 
%
\subsubsection{Block Matching and Optical Flow}
Block matching involves taking a patch of image in a frame and searching for the best match in a spatial window in the next time frame. \citet{langeland2005experimental} implemented this on 2D echo RF (radio frequency) images. \citet{lubinski1999speckle} also implemented this on RF images and further refined displacement estimation in the beam direction using zero-crossings of the phase of the complex correlation function. Optical flow methods assume that the intensity of a point in a moving image is consistent across time and that motion is responsible for temporal intensity variation. \citet{song1991computation} applied this to model cardiac motion in 3D CT images. These methods can be time-consuming due to a large search space and also lack a regularization term. 
%
%
\subsection{Shape Matching/Tracking Methods}
%
Shape-based methods try to match shape/image descriptors derived from a point set. Pre-processing is necessary to generate the points either by simple edge/feature detection algorithms or by a more sophisticated segmentation algorithm. Post-processing is also generally required for smoothing and dense field generation as the solutions are sparse \citep{papademetris2002estimation}.
\subsubsection{Frame-to-frame Matching}
\citet{chui2003new} proposed a point matching algorithm that modeled deformation using non-rigid thin plate spline parameterization and used it to align point sets derived from brain-imaging. The correspondences that map point sets are fuzzy (non-binary) initially and are refined iteratively to obtain one-to-one binary correspondences. 
\citet{belongie2000shape} introduced the shape context feature, which is more global than the local curvature, and used it to match point sets. They solve a weighted bipartite graph matching problem using the Hungarian algorithm to obtain one-to-one correspondences. 
%
%
\subsubsection{Temporal Tracking}
We previously proposed a method that tracks individual points on myocardial surfaces through time \citep{parajuli2016integrated}. Points on the myocardial surfaces form nodes in a graph and edges exist between points and their spatial neighbors in the next time frame. The motion of an individual point is then posed as the shortest path through this graph. Our current work improves this by modeling the motion of all points on the surface together as opposed to individually. \citet{berclaz2011multiple} used a flow network structure to build a fully spatiotemporal model for an object tracking problem. We expand upon their work, as will be seen below, by providing a probabilistic mechanism of outlier handling and by accounting for periodic motion. Furthermore, while their work handles the uncertainty in the nodes of the graph, we handle uncertainty in the edges to solve for correspondences.
%
%
\subsection{Post hoc Regularization Models}
Methods lacking in inherent regularization, or producing a sparse set of displacements as our method does, rely on post hoc regularization of the initial tracking results to produce smooth displacement fields. \citet{papademetris2002estimation} first estimated initial correspondences between myocardial surfaces using a shape matching approach. The initial estimation was then regularized by using a biomechanically inspired finite element method approach.
\par \citet{compas2014radial} et al. proposed the use of radial basis functions to generate smooth and dense displacements from the integration of sparse sets of shape and speckle tracking displacements in. We expanded this strategy to impose further smoothness and biomechanical constraints on the displacement fields in \citet{parajuli2015sparsity}. \citet{lu2017learning} learned how to regularize noisy motion by training a neural network to filter noisy 4D Lagrangian displacement vector fields. 
%
%
\section{Methods}
\begin{figure}
	\centering
	\includegraphics[scale=.32]{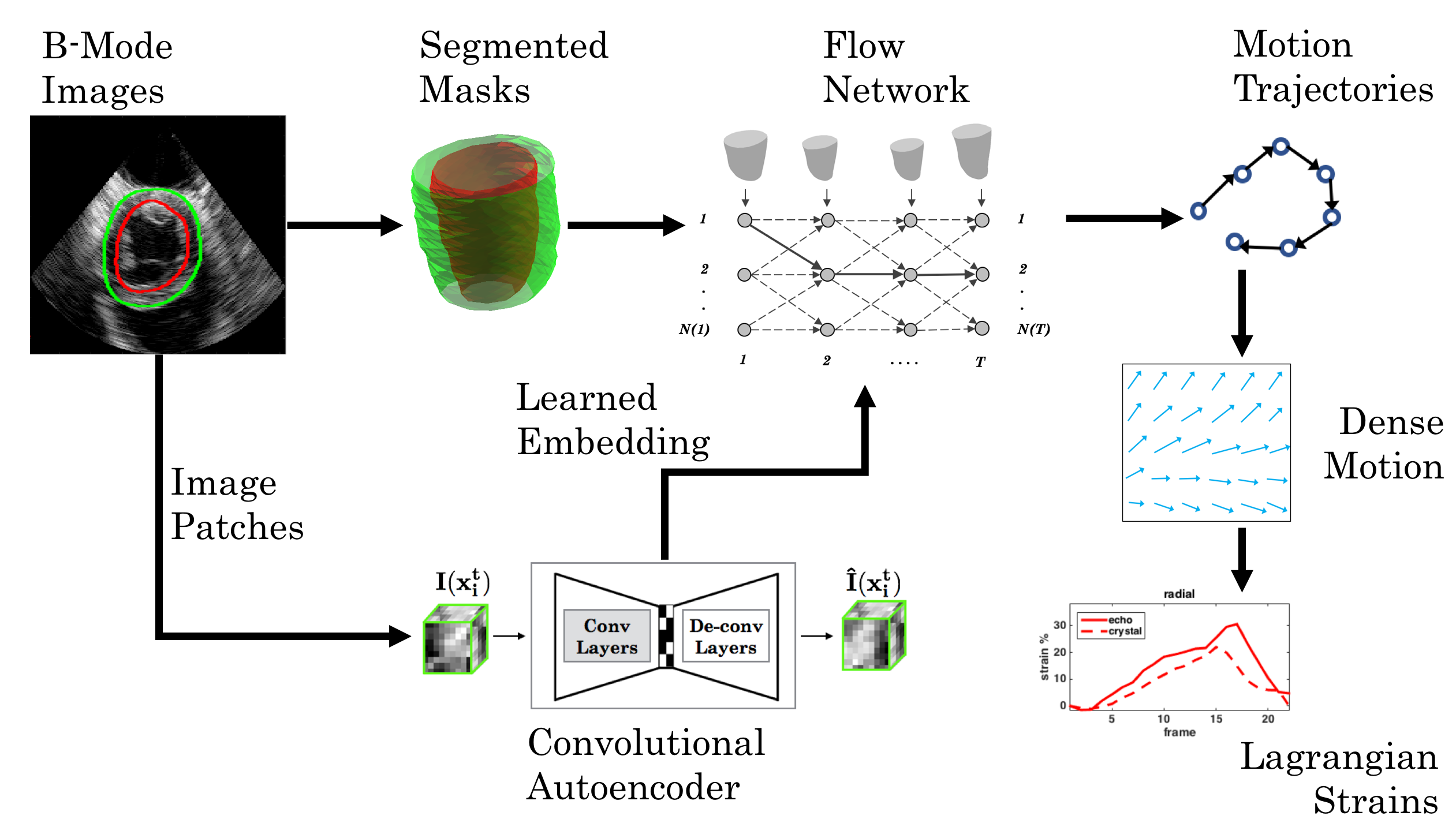}
	\caption{Overall method outline.}
	\label{fig:system_pipeline}
\end{figure}
\par Many point matching methods that model rigid or non-rigid deformations try to achieve one-to-one or symmetric matches \citep{belongie2000shape, chui2003new}. Even within the free-form deformation framework, diffeomorphic transformations are a popular choice, which utilize a similar concept. The resulting displacement fields from such mappings are more realistic and robust to noise and artifacts. 
However, matching algorithms that impose one-to-one correspondence only at a frame-to-frame level cannot guarantee that a composition of those one-to-one correspondences is also similarly one-to-one. Thus, in the work presented in this paper below, we develop an approach that i.) incorporates global one-to-one correspondence for all points being tracked and ii.) accounts for the cyclical nature of cardiac motion.

\begin{figure}
	\centering
	\includegraphics[scale=.28]{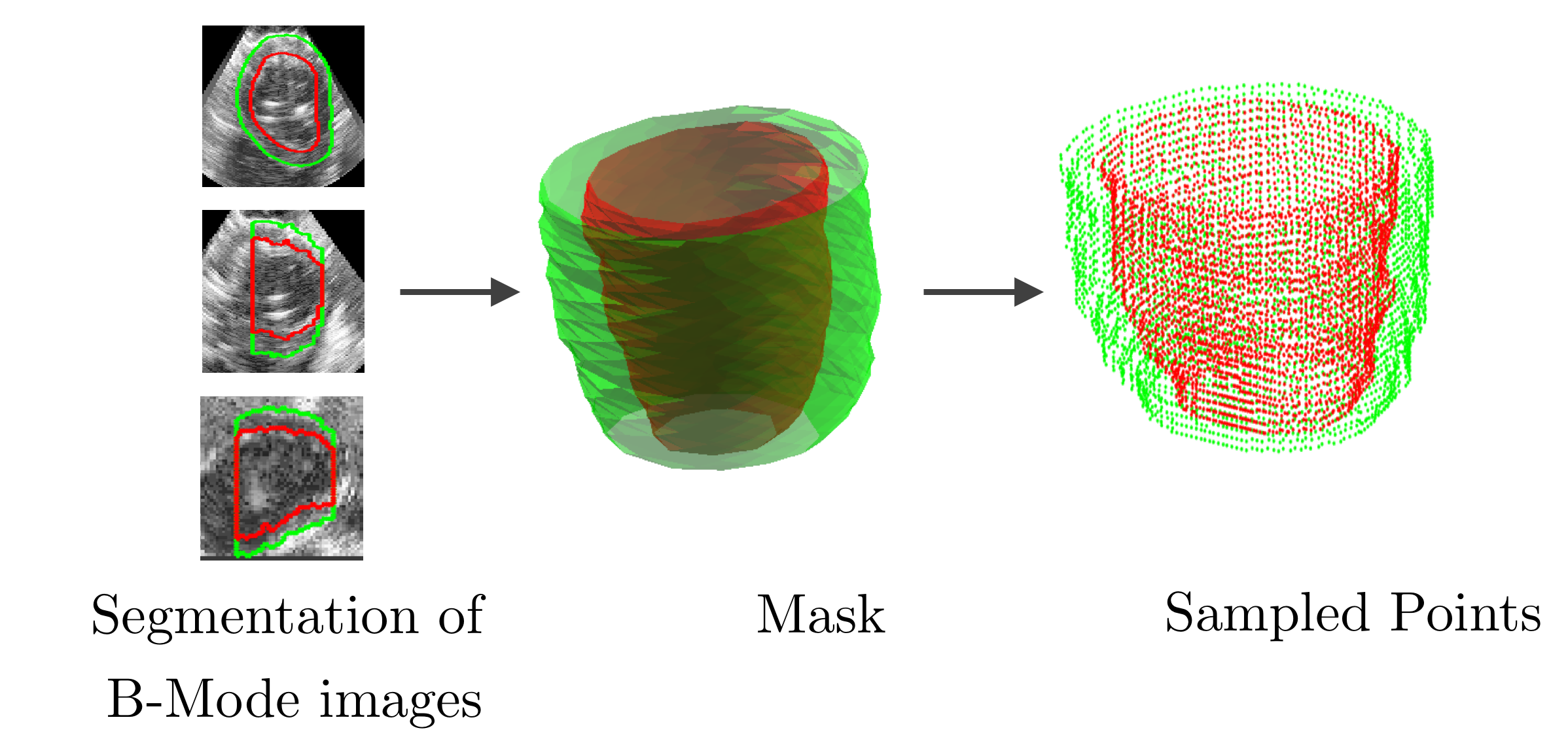}
	\caption{Preprocessing steps for \textit{in vivo} data to get endocardial and epicardial surface points.}
	\label{fig:preprocess_surfaces}
\end{figure}
%
\subsection{Constrained Flow Optimization}
\par
\begin{figure}
	\centering
	\includegraphics[scale = .43]{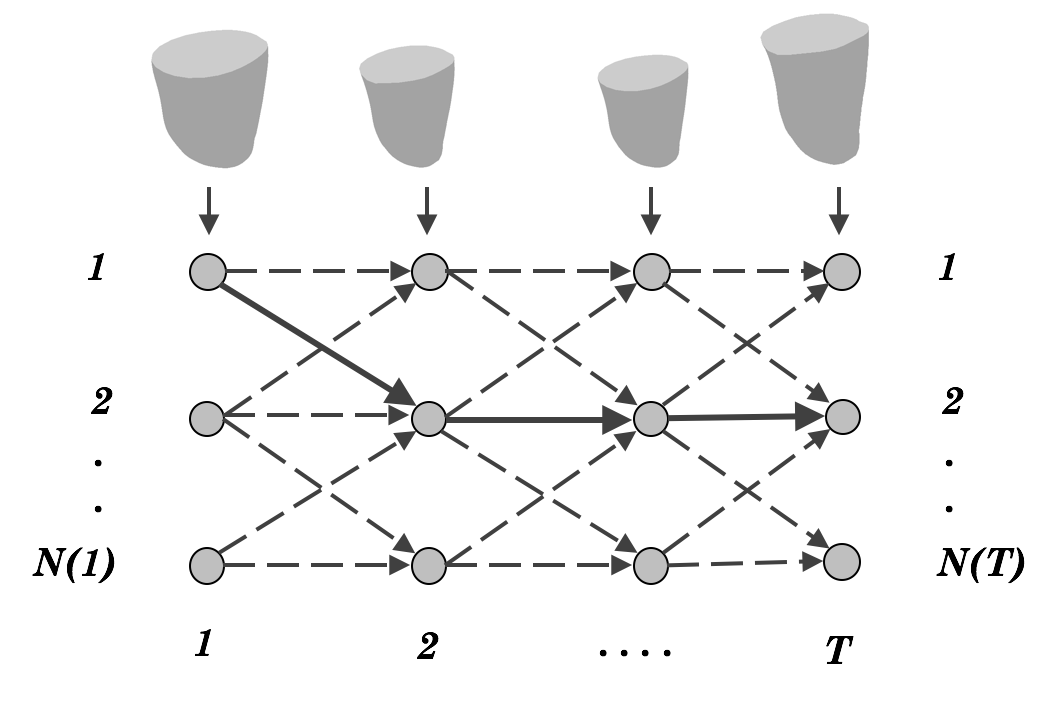}		
	\caption{Nodes, edges and other relationships in the network. The point sets are sampled from the myocardial surface sequence at each frame.}
	\label{fig:graphical_structure}
\end{figure}

 First, point clouds are obtained by uniformly sampling the endocardial and epicardial surfaces (see figure \ref{fig:preprocess_surfaces}). The sequence of point clouds, through the cardiac cycle, is then set up as nodes in a graph with directed edges between points and their match candidates in the next time frame. Each node is endowed with a feature vector representing local appearance characteristics and the match candidates are chosen based on feature distances between nodes and their spatial neighbors. This is illustrated in Figure \ref{fig:graphical_structure}.
\begin{table}
	\centering
	\caption{Notations}
	\begin{tabular}{r  l}
		\hline
		$T$ & $\text{Number of frames} $ \\
		$N(t)$ & $\text{Number of points per frame}, $ \\ 
		&\qquad \qquad $t \in [1:T] $ \\
		$x^t_i$ & $i^{th} \text{ point of frame } t,$ \\ 
		& \qquad \qquad $i \in [1:N(t)]$  \\
		$e^t_{ij}$ & \text{Edge from point } i \text{ in frame } t \\
		& $\text{ to point } j \text{ in frame } t+1$  \\
		& \qquad \qquad $i \in [1:N(t)]$  \\
		& \qquad \qquad $j \in [1:N(t+1)]$  \\
		$w^t_{ij}$ & $\text{Weight associated with } e^t_{ij}$  \\ 
		$f^t_{ij}$ & $\text{Flow through } e^t_{ij}$  \\ 
		$ \eta (t, i)$ & $\text{Indices of points in the}$ \\
		& $\text{neighborhood of } x^t_i \text{ in frame } t+1$ \\
		\hline
	\end{tabular}
\end{table}
\par The edges capture particle (tissue) motion possibilities, and their weights capture the likelihood of the motion. We have $T$ time frames in total with $N(t)$ ($t \in [1:T]$) points per frame. Each node is defined as $x^t_i$ ($i \in [1:N(t)]$), where an edge $e^t_{ij}$ exists between $x^t_i$ at time $t$ and its neighbor $x^{t+1}_j$ (based on feature distances) at time $t+1$ ($i \in [1:N(t)]$ and $j \in [1:N(t+1)]$). The flow through an edge $e^t_{ij}$ in this network is captured by the binary-valued variable $f^t_{ij}$, and the corresponding edge weight is $w^t_{ij}$. $f^t_{ij} = 1$ implies that the point $x^t_{i}$ and $x^{t+1}_j$ are a match.
\par
\begin{figure}
	\centering
	\subfloat[A node $x^t_i$ and outgoing flows $f^t_{ij}$, which must sum to 1]{\includegraphics[scale=.43]{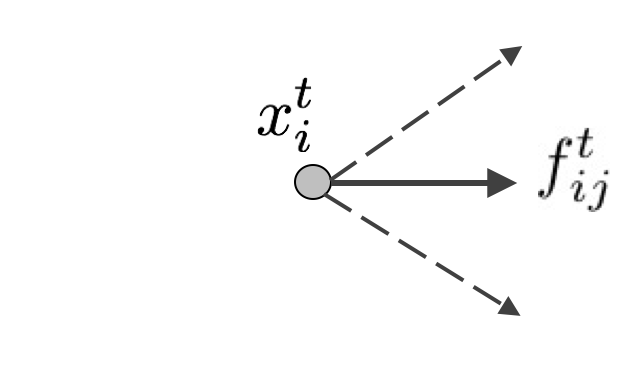}} 
	\qquad 
	\subfloat[A node $x^t_i$ and outgoing flows $f^t_{ij}$ and incoming flows $f^t_{ij}$. These must be in balance.]{\includegraphics[scale=.43]{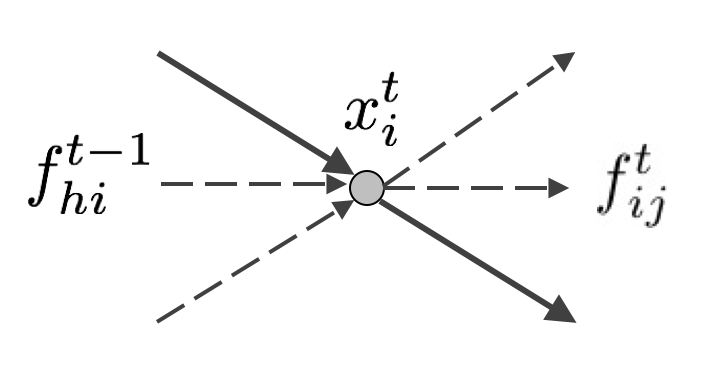}}
	\caption{A node and different edges/flows visualized}
	\label{fig:node_constraints}
\end{figure}

 We would like to solve for flow $f$ that is proportional to the edge weights $w$. This amounts to maximizing the inner product $w'f$, subject to the following constraints at each node $x^t_i$ ($\eta(t, i)$ indexes neighbors of $x^t_i$ in frame $t + 1$):
\begin{enumerate}
	\item Flows are non-negative:
	\begin{equation}
	\forall t, i \qquad f \geq 0.
	\end{equation}
	\item Sum of outgoing flows is less than or equal to one ($C_{out}$, see Figure \ref{fig:node_constraints}a).
	\begin{equation}
	\label{eqn:cout_sum}
	\forall t, i \qquad \sum_{j\in \eta(t, i)} f^t_{ij} \leqslant 1
	\end{equation}
	\item Sum of outgoing and incoming flows is equal ($C_{bal}$, see Figure \ref{fig:node_constraints}b).
	\begin{equation}
	\forall t, i \sum_{h:i\in \eta(t-1, h)} f^{t-1}_{hi} = \sum_{j\in \eta(t, i)} f^{t}_{ij}
	\end{equation}
\end{enumerate}
%
%
\par $C_{out}$ ensures that total flow is preserved from one frame to another. By itself, it would result in complete trajectories that traverse through the best possible path but without any consideration for spatial consistency with other trajectories. $C_{bal}$ is enforced so that incoming and outgoing flows through nodes are equal. This is helpful in preventing many-to-one correspondences, which create an uncharacteristic stretching and shrinking in the displacement fields. One-to-many correspondences are also avoided for the same reason. 
\par Even though we would ideally like to solve an integer programming (IP) optimization to get a binary-valued solution for $f$, we start with a linear programming (LP) relaxation, and solve the following optimization instead (all variables are considered in their vectorized forms here for simplicity):
\begin{equation}
	\centering
	\label{eqn:strict_optim}
	\begin{aligned}
		&\text{Maximize}  &&  w' f\\
		&\text{subject to} && f \geqslant 0, &&   C_{out} f \leqslant 1, &&   C_{bal} f \leqslant 0 \\
	\end{aligned}
\end{equation}
\par Despite the relaxation of an IP into an LP, we still obtain a binary-valued solution. This is of great consequence because solving IPs directly is an NP-hard problem. Due to the special nature of the constraint matrices that define our LP, our solutions remain integer-valued. This builds upon work that proves that inequalities in LP with constraint matrices that satisfy the total unimodularity property, and that have an integral right-hand side, lead to an integral solution \citep{hoffman2010integral}. A unimodular matrix is a square integer matrix with determinant $+1$ or $-1$. This results in the inverse of these matrices also being an integer matrix. \citet{berclaz2011multiple} show that such flow balance constraint matrices satisfy total unimodularity. This allows us to obtain fully connected non-intersecting trajectories throughout the cardiac cycle. 
\par
\begin{figure}
	\centering
	\subfloat[$C_{out}$ only: outgoing flows sum to $\leq 1$ leading to shortest paths tracking for individual points.]{\includegraphics[scale = .32]{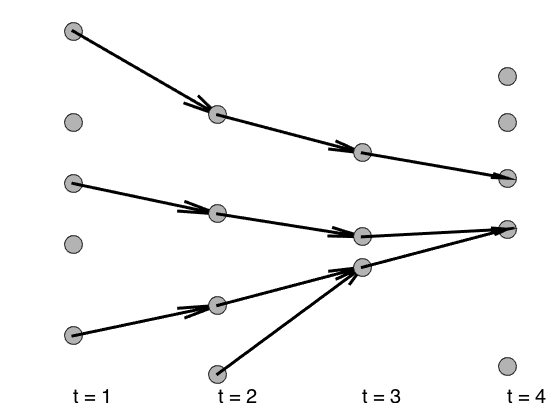}} \qquad			
	\subfloat[$C_{out}$ and $C_{in}$: outgoing and incoming flows sum to $\leq 1$ at all nodes leading to frame-to-frame maximum bipartite match.]{\includegraphics[scale = .32]{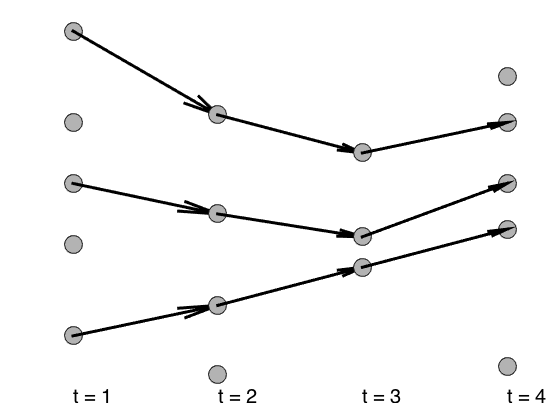}} \\
	\subfloat[$C_{out}$ and $C_{bal}$: outgoing and incoming flows sum to $\leq 1$ and are also equal at all nodes leading to total maximum bipartite match.]{\includegraphics[scale = .32]{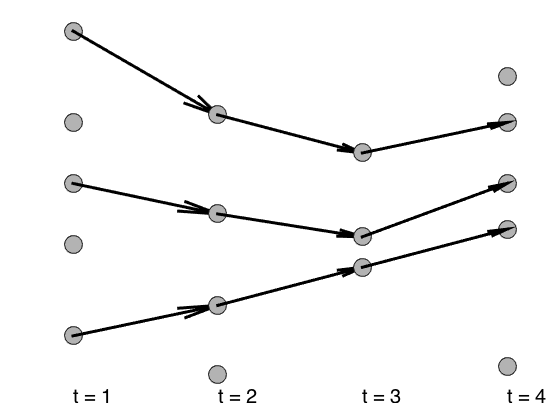}} \qquad			
	\subfloat[$C_{out}$, $C_{bal}$ and $C_{loop}$: additional equality (balance) constraints lead to a flow balance between first and last nodes leading closed-looped behavior as well.]{\includegraphics[scale = .32]{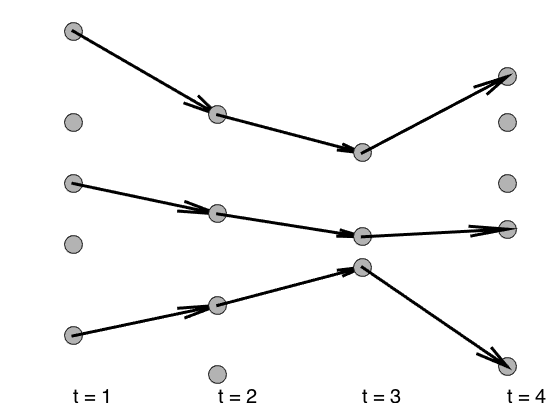}}	
	\caption{Outcomes of applying different constraints on 1D+t point sets, with points stacked vertically in space and horizontally in time.}
	\label{fig:constraint_outcomes}
\end{figure}

 It is instructive to think of the effects of the constraints in terms of an equivalent graph-based matching/tracking methods. For instance, imposing only $C_{out}$ results in shortest path tracking for each point. In addition to $C_{out}$, if we also impose $C_{in}$, such that the sum of incoming flows at a node is also constrained ($C_{in} f \leq 1$), we achieve a maximum bipartite match, with one-to-one correspondence, between point sets in consecutive time frames. 
Finally, if $C_{out}$ and $C_{bal}$ are both enforced, we obtain a complete maximum bipartite match, which results in one-to-one correspondence that extends through the entire cardiac cycle ($C_{in}$ gets implicitly imposed in this case). The effects of these constraints on the resulting trajectories are illustrated in figure \ref{fig:constraint_outcomes}(a)-\ref{fig:constraint_outcomes}(c).
%
\par Since our constraint matrices are very sparse, the LP can be solved very efficiently. We used the CVX package (in MATLAB), for specifying the LP and other optimizations in this work \citep{cvx, gb08}. CVX was used in conjunction with the MOSEK solver for solving the LP, which uses the interior-point method \citep{mosek}.
%
\subsubsection{Constraint Matrices}
\begin{figure}
	\centering
	\includegraphics[scale = .46]{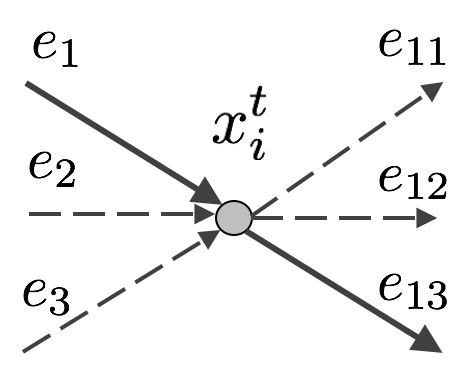}		
	\caption{Incoming and outgoing edges at a node.}
	\label{fig:out_balance_fig}
\end{figure} 
At a node associated with $x_i^t$, let $e_1, e_2, e_3$ be the incoming edges and let $e_{11}, e_{12}, e_{13}$ be the outgoing edges (in vectorized form). Fig \ref{fig:out_balance_fig} illustrates this relationship and the form of the equation corresponding to a row of the $C_{bal}$ matrix appears as follows:
\begin{equation}
	\begin{bmatrix}
	-1 -1 -1 & \ldots \quad 1 \quad 1 \quad 1
	\end{bmatrix}
	\begin{bmatrix}
	f_1 \\ f_2 \\ f_3  \\ \vdots \\ f_{11} \\ f_{12} \\ f_{13} 
	\end{bmatrix} \leq 0.
	\label{eqn:cbal_eqn}
\end{equation}
%
%
Similarly, $C_{out}$ has $1$'s where $C_{bal}$ also has $1$ and zeros elsewhere. Conversely, $C_{in}$ has $1$'s where $C_{bal}$ has $-1$.
\par Fig \ref{fig:constraint_outcomes} displays how imposing $C_{out}$, $C_{out}$ and $C_{in}$, $C_{out}$ and $C_{bal}$ affects tracking outcomes in a toy $1D+t$ problem. The points are stacked vertically for each time step (total of $5$). We can clearly see how just imposing $C_{out}$ leads to the qualitatively worst result in Fig \ref{fig:constraint_outcomes}(a). This is because there is nothing preventing two trajectories from merging and occupying the same nodes. 
\par We can also notice, in figure \ref{fig:constraint_outcomes}(c), as $C_{bal}$ is imposed strictly, one-to-one point correspondence leads to non-overlapping complete trajectories. However, nodes on the top areas in figure \ref{fig:constraint_outcomes}(c) have no trajectories passing through. In an earlier implementation of this algorithm, we attempted to remedy situations like this by relaxing this strict balance constraint \citep{parajuli2017flow}. In addition to that, we also added spatiotemporal smoothness constraints. However, in this work, we impose $C_{bal}$ strictly and develop a different strategy of regularization suitable to cardiac motion. We discuss this strategy next.
%
\subsubsection{Imposing Loop Constraints}
\begin{figure}
	\centering
	\includegraphics[scale=.46]{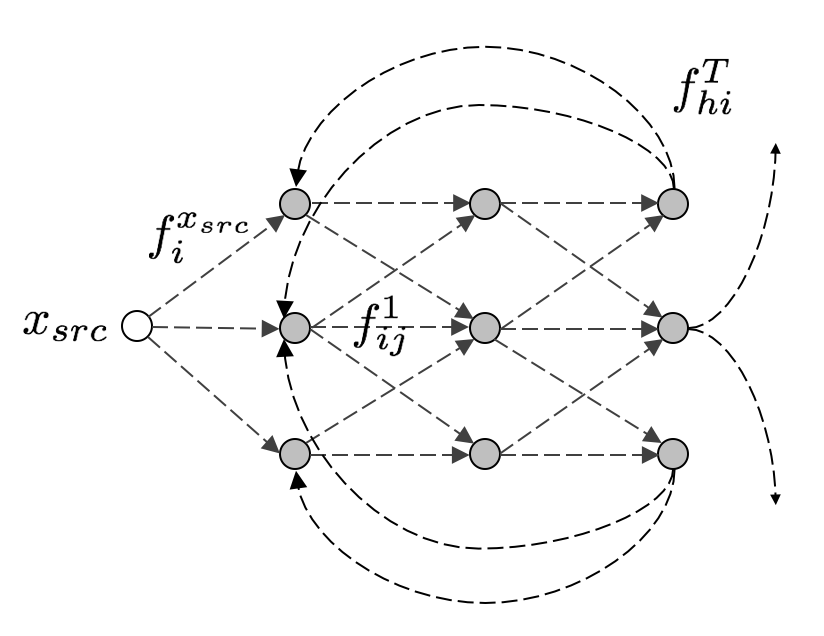}
	\caption{A simple flow network displaying the additional loop edges between the last frame and the first (not all shown) which helps us obtain closed-looped trajectories. The source node and edges are also shown.}
	\label{fig:closed_looped}
\end{figure}
A simple extension to the algorithm described so far encourages trajectories to end up in close proximity to their start location. This is done by adding one more set of edges (and constraints) between the points in the last frame and the first frame (see figure \ref{fig:closed_looped}.). First, we impose the following constraint between the flow from source node $x_{src}$ (let the flow be $f^{x_{src}}_i$) and the flow from the first frame to second ($f^1_{ij}$):
\begin{equation}
	\label{eqn:loop_constraint_1}
	\forall i, i \in [1:N(1)], \qquad f^{x_{src}}_i = \sum_{j \in \eta(1, i)} f^1_{ij}.
\end{equation}
Here, since each node in frame $1$ is connected directly to $x^{src}$, we are saying that if there is a flow into that node from the source, there has to be a flow out of the node going into frame $2$ as well. Next, we impose a balance between flow from the last frame to the first frame ($f^T_{hi}$ via the loop edges) and flow from first to second ($f^1_{ij}$):
\begin{equation}
	\label{eqn:loop_constraint_2}
	\forall i, i \in [1:N(1)], \qquad \sum_{h : i \in \eta(T, h)} f^T_{hi} = \sum_{j \in \eta(1, i)} f^1_{ij}.
\end{equation}
This is the key flow balance constraint that encourages trajectories to be closed-looped. Since edges always exist between nodes that are close spatially, and we already have a mechanism for obtaining non-overlapping complete trajectories, this helps us ensure that our trajectories are roughly closed-looped. Finally, to make sure that the total flow leaving the source node is also equal to the total flow leaving the last frame (and going to the first implicitly), we also impose the following constraint:
\begin{equation}
	\label{eqn:loop_constraint_3}
	\sum^{N(1)}_{i=1} f^{x_{src}}_i = \sum^{N(1)}_{i=1} \sum_{h : i \in \eta(T, h)} f^T_{hi}.
\end{equation}
\par We shall call these constraints $C_{loop}$, which is applied in the same way as $C_{bal}$ and is incorporated within it during the optimization (see equation \ref{eqn:strict_optim}). 
In figure \ref{fig:constraint_outcomes}(d), we can see how the loop constraint helps us recover trajectories that lead to the end points of the trajectories being in close proximity of their start points, which is what we expect due to the periodicity of the LV motion/displacement.
%
%
%
\subsubsection{Outlier Handling}
\begin{figure}
	\centering
	\includegraphics[scale=.32]{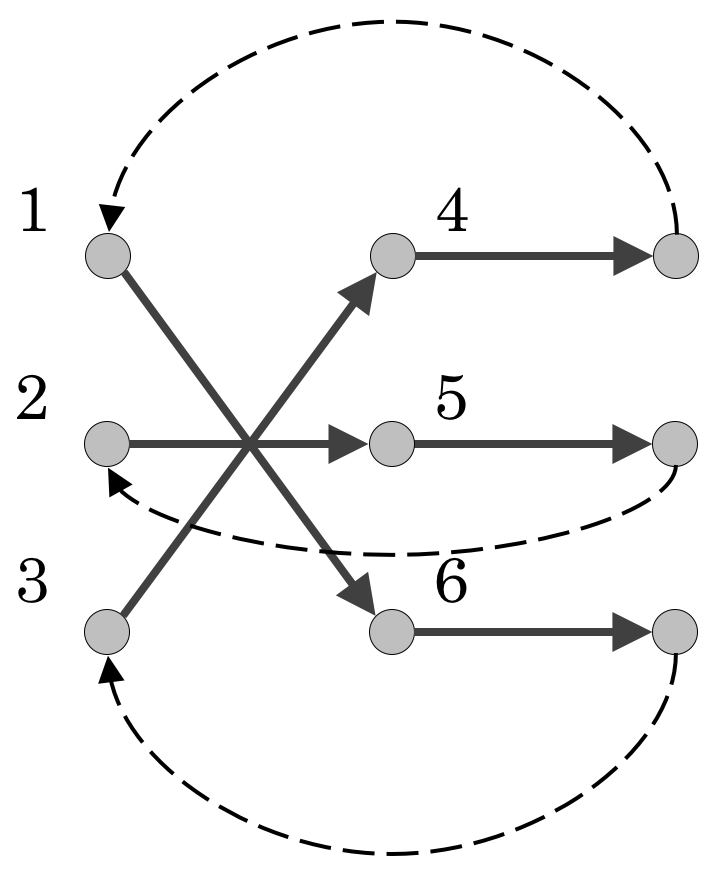}
	\caption{An unlikely, but a valid scenario where balance and closed loop constraints are satisfied but the result is poor qualitatively.}
	\label{fig:bad_tracking_conceptual}
\end{figure}

\begin{figure}
	\centering
	\subfloat[Trajectories computed from random $1D+t$ data.]{\includegraphics[scale = .28]{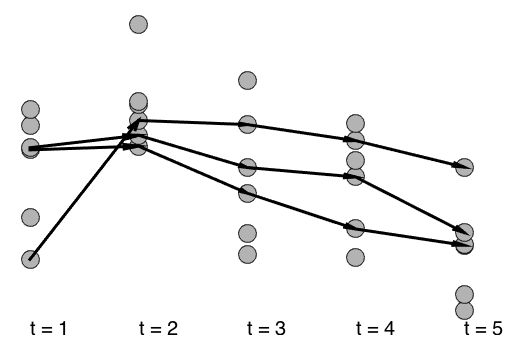}} \qquad 		
	\subfloat[Trajectories computed from random $1D+t$ data after thresholding.]{\includegraphics[scale = .28]{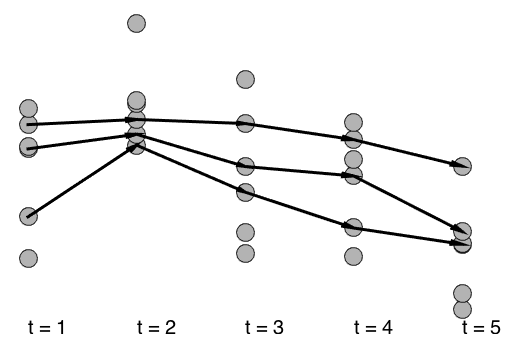}} 
	\caption{Outcome of applying thresholding (of $P_{th} = .3$) on the edge weights.}
	\label{fig:degenerate_example}
\end{figure}

Because we are maximizing the flow through the network, our algorithm always solves for the maximum number of possible trajectories, even if they are of very low quality. Figure \ref{fig:bad_tracking_conceptual} displays such an example. The ideal match would have been if point $1$ was matched to point $4$ and point $3$ to $6$, but instead, the opposite has happened. We tackle this by probabilistic thresholding. All edges with weights below a certain threshold $P_{th}$ are omitted. An example is shown in figure \ref{fig:degenerate_example} with randomly generated points for illustration.
%
%
\subsubsection{Edge weight calculation.}
Each edge $e^t_{ij}$ has the following weight:
\begin{equation} 
	\label{eqn:edge_wt}
	w^t_{ij} = exp\left( \frac{-||x^t_i - x^{t+1}_{j}||^2}{2\sigma^2_x} \right) exp\left( \frac{-|| F(x^t_i) - F (x^{t+1}_{j}) ||^2}{2\sigma^2_{f}} \right)
\end{equation}
\par $F$ can be any shape or appearance-based feature associated with $x^t_i$ and $x^{t+1}_j$. $\sigma_x$ and $\sigma_{f}$ are normalization constants and are calculated using the standard deviations of Euclidean and feature distances for each image frame.
%
%
%
\subsection{Neural Network Based Appearance Features}
\par
\begin{figure}
	\centering
	\includegraphics[scale = .28]{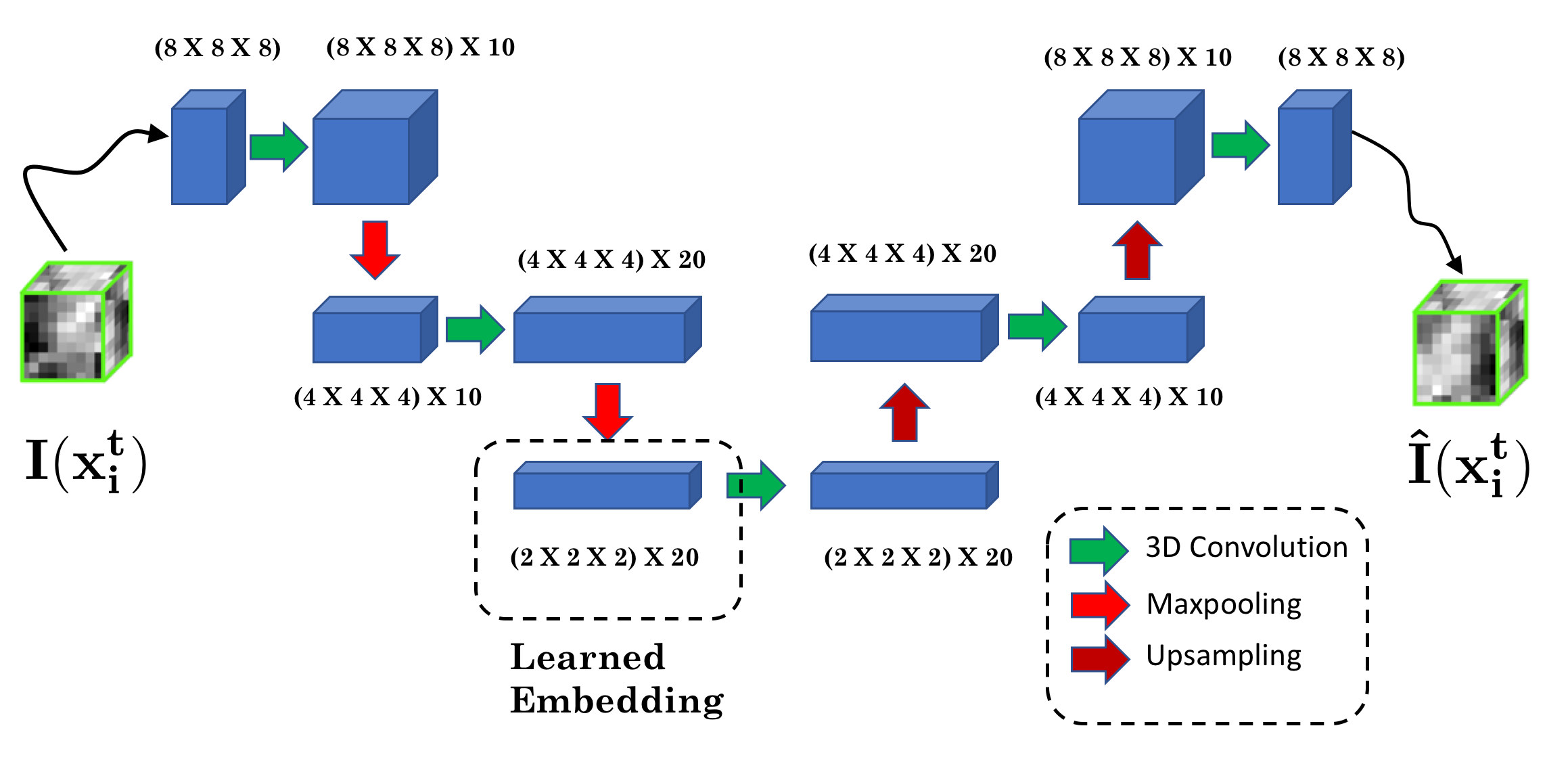}		
	\caption{Autoencoder network: A low dimensional embedding of image patches that captures images statistics is learned.}
	\label{fig:unsupervised_network}
\end{figure}

 In contrast to our previous work \citep{parajuli2017flow}, we use a convolutional autoencoder to learn appearance features because we do not have ground truth trajectories for \textit{in vivo} data. Convolutional autoencoders are known to produce a state-of-the-art solutions in unsupervised learning problems. The structure of the network is shown in figure \ref{fig:unsupervised_network} and has the standard encoder-decoder format. We use Euclidean distance to quantify the similarity between two learned embeddings. The learned embedding $F$ is used to calculate edge weights in equation \ref{eqn:edge_wt}.
%
\subsection{Dense Field Generation}
\par We use radial basis functions (RBFs) as interpolants and calculate dense displacement fields, which is necessary for generating dense motion trajectories and Lagrangian strain which helps us assess myocardial function. This is based on our previous work in \citet{parajuli2015sparsity} and that of \citet{compas2014radial}.  
\par We regularize our motion field ($U$) by imposing sparsity on the weights associated with the basis functions representing our motion field that account for tissue incompressibility. We do so under the assumption that the cardiac tissue is roughly incompressible and therefore, the motion vector field should be roughly divergence-free (i.e., $\nabla \cdot U = 0$) \citep{song1991computation}. Finally, we also mildly penalize the norm of the spatial derivatives ($\nabla U$) to discourage jumps and discontinuities in the motion fields. We use a compactly supported basis function, which results in a sparse basis matrix unlike other popular choices such as Gaussian and thin plate spline based RBFs. Sparse matrices are more conducive to numerical optimization \citep{compas2014radial, wendland1995piecewise}. Once dense displacements are obtained, Lagrangian strain is calculated using the method described in \citet{yan2007boundary}. 
\section{Experiments and Results}
\par We explored how changes in key parameters affect performance via a synthetic dataset where ground truth was available. Although the synthetic data do not fully capture all the variation present in vivo echocardiographic images, it helped us establish best practices. The autoencoder-derived features are generated using models trained in a leave-one-out fashion where, for every synthetic data sequence, a different model was trained without using that sequence.
For the in-vivo data, we obtained endocardial and epicardial surfaces using an automated (except the first frame) level-set segmentation method \citep{huang2014contour}. In the synthetic case, segmentation masks were generated from the ground truth mesh masks, hence a segmentation step was not required during pre-processing. Other than that, the steps are the same for the in-vivo data as well. As a reminder, the overall method is depicted in figure \ref{fig:system_pipeline}.
\subsection{Synthetic Data}
%
\subsubsection{Data Description}
\par
\begin{figure}
	\centering
	\subfloat[Short-axis view]{\includegraphics[scale=.4]{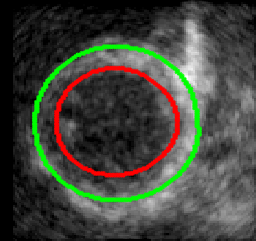}}
	\qquad \qquad 
	\subfloat[Long-axis view]{\includegraphics[scale=.26]{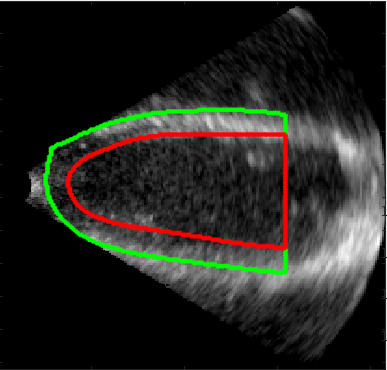}}
	\caption{Synthetic data image example with endocardial and epicardial contours (normal data).}
	\label{fig:leuven_example}
\end{figure}

 We used 8 synthetic 3D+t ultrasound image sequences developed by \citet{alessandrini2015pipeline}. The dataset consisted of 3 categories of image sequences. The first consisted of just one normal sequence (normal). The second consisted of ischemic sequences with ischemia in the distal and proximal left anterior descending artery (LADDIST and LADPROX), right circumflex artery (RCA) and left circumflex artery (LCX). The third consisted of dilated myocardium sequences - one synchronous (SYNC) and two dyssynchronous, induced by left bundle branch block (LBBB and LBBBSMALL). Examples of the synthetic data - 2D slices from one image sequence at end diastole and corresponding contours - are given in figure \ref{fig:leuven_example}.
%
%
%
\par  
We report tracking errors on 2250 myocardial mesh points, for which the positions through the sequence were provided as ground truth. These points were evenly distributed in the endocardium, epicardium and in the mid-wall. We calculated distances between ground truth mesh points and mesh points from our tracking algorithm by propagating the known first frame mesh points through time. The errors are summarized using median and interquartile range (IQR). Overall errors, aggregated over all time frames, are displayed separately from end diastolic (last frame) and ES errors because the number of data points is different. Analyzing errors at ES and ED is important as many clinical readings are made at these time points.

%
\subsubsection{Parameter Selection for FNT}
%
\par
\begin{figure}
	\centering
	\subfloat[Along the circumference of the surfaces, $\theta_{fr}$ points are uniformly sampled in terms of angle.]{\includegraphics[scale=.28]{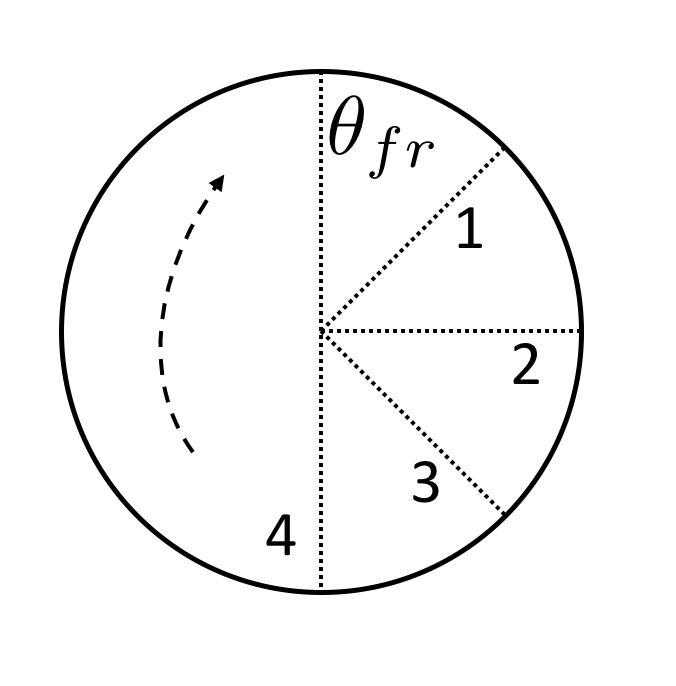}}
	\qquad \qquad
	\subfloat[Along the long (z) axis of the surface, $Z_{fr}$ points are uniformly sampled. ]{\includegraphics[scale=.28]{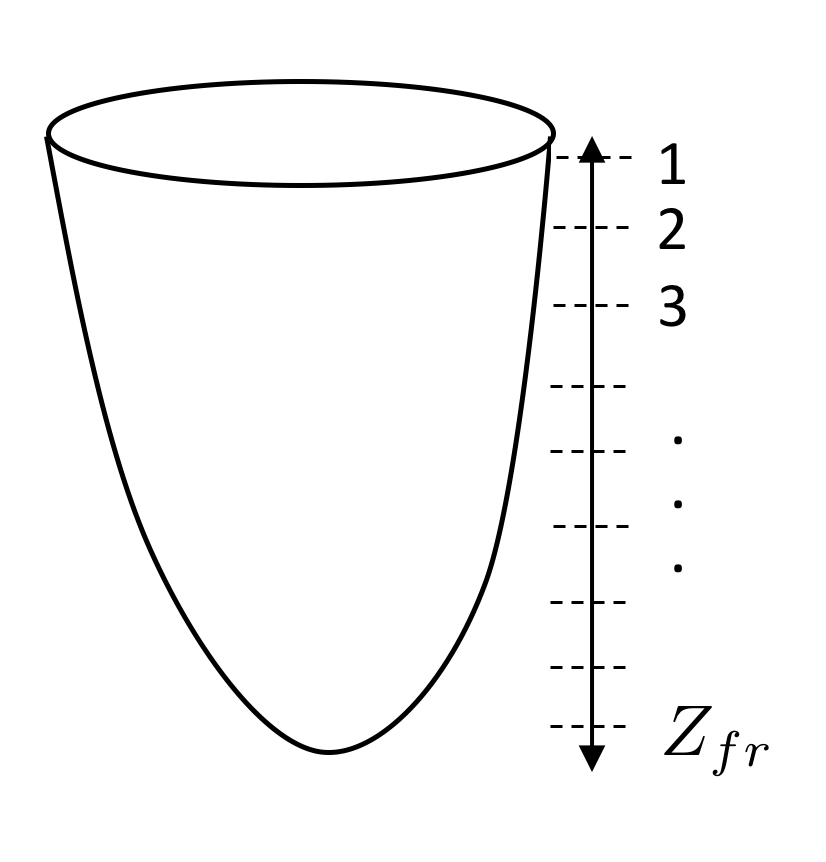}}
	\caption{Sampling scheme.}
	\label{fig:angular_sampling}
\end{figure}

 We vary $Z_{fr}$, $\theta_{fr}$ and $NK$, which control the level of sampling  of our surface masks and graph neighbor assignment (see table \ref{table:param_consider} for description). We adopted a cylindrical sampling strategy where we sample uniformly along the $z$ axis and along the circumference (see figure \ref{fig:angular_sampling}). We present the median tracking errors (MTE) under different combination of these parameters on the normal data in table \ref{table:changing_z_theta}. The combination of $Z_{fr} = 40$, $\theta_{fr} = 30$ and $NK = 3$ provided the lowest overall MTE on the normal data. 
\begin{table}
	\centering
	\caption{Parameters that were tuned for the FNT algorithm.}
	\begin{tabular}{|p{2cm}|p{7cm}|}
		\hline 
		\textbf{Name} & \textbf{Description} \\
		\hline 
		$NK$ & Number of nearest neighbors (by feature distance) in consideration for next frame. \\
		\hline 
		$Z_{fr}$ & Number of slices sampled in the long ($z$) axis  per frame. \\
		\hline 
		$\theta _{fr}$ & Angular sampling along the circumference. (roughly along the short axis.)\\
		\hline 
		$P_{th}$ & Probabilistic threshold for outlier edges removal. \\
		\hline
	\end{tabular}
	\label{table:param_consider}
\end{table}

\begin{table}
	\centering
	\caption{Result of changing $Z_{fr}$, $\theta_{fr}$ and $NK$ on the normal data and median square errors (MTE).}
	\begin{tabular}{|c|c|c|c|c|c|}
		\hline
		$\mathbf{Z_{fr}}$ & $\mathbf{\theta_{fr}}$ & $\mathbf{NK}$ & \textbf{Overall/mm} & \textbf{ES/mm} & \textbf{ED/mm}\\ \hline 
		30 & 15 & 3  & 0.96 $\pm$ 0.74 & 1.13 $\pm$ 0.75  & 1.01 $\pm$ 0.82  \\ \hline 
		30 & 30 & 3  & 0.94 $\pm$ 0.71 & 1.25 $\pm$ 0.87  &  0.77 $\pm$ 0.56 \\ \hline 
		30 & 40 & 3  & 0.95 $\pm$ 0.73 & 1.30 $\pm$ 0.87 & 0.82 $\pm$ 0.53  \\ \hline 
		30 & 15 & 5  & 0.90 $\pm$ 0.71 & \textbf{1.09 $\pm$ 0.68}  &  0.90 $\pm$ 0.79 \\ \hline 
		30 & 30 & 5  & 0.97 $\pm$ 0.73 & 1.24 $\pm$ 0.83  & 0.98 $\pm$ 0.77 \\ \hline 
		30 & 40 & 5  & 0.93 $\pm$ 0.70 & 1.20 $\pm$ 0.80  & 0.90 $\pm$ 0.65  \\ \hline \hline 
		40 & 15 & 3  & 0.90 $\pm$ 0.68 & 1.17 $\pm$ 0.76 & 0.78 $\pm$ 0.66 \\ \hline 
		40 & 30 & 3  & \textbf{0.86 $\pm$ 0.65} & 1.09 $\pm$ 0.73 & 0.82 $\pm$ 0.58 \\ \hline 
		40 & 40 & 3  & 0.90 $\pm$ 0.66 & 1.21 $\pm$ 0.73 & 0.77 $\pm$ 0.53  \\ \hline 
		40 & 15 & 5  & 0.91 $\pm$ 0.66 & 1.11 $\pm$ 0.71 & 0.87 $\pm$ 0.64  \\ \hline 
		40 & 30 & 5  & 0.95 $\pm$ 0.74 & 1.24 $\pm$ 0.75 & 0.87 $\pm$ 0.75  \\ \hline 
		40 & 40 & 5  & 0.95 $\pm$ 0.71 & 1.16 $\pm$ 0.73 &  0.92 $\pm$ 0.79 \\ \hline \hline 
		50 & 15 & 3  & 0.96 $\pm$ 0.75 & 1.15 $\pm$ 0.72  & 1.02 $\pm$ 0.84 \\ \hline 
		50 & 30 & 3  & 0.93 $\pm$ 0.74 & 1.21 $\pm$ 0.79  & 0.86 $\pm$ 0.78 \\ \hline 
		50 & 40 & 3  & 0.91 $\pm$ 0.72 & 1.09 $\pm$ 0.72  & 0.80 $\pm$ 0.72  \\ \hline 
		50 & 15 & 5  & 0.90 $\pm$ 0.67 & 1.13 $\pm$ 0.82 & 0.86 $\pm$ 0.59  \\ \hline 
		50 & 30 & 5  & 0.88 $\pm$ 0.73 & 1.31 $\pm$ 0.75 &  \textbf{0.75 $\pm$ 0.67} \\ \hline 
		50 & 40 & 5  & 0.91 $\pm$ 0.75 & 1.34 $\pm$ 0.80 & 0.83 $\pm$ 0.66 \\ \hline  
		\end{tabular}
	\label{table:changing_z_theta}
\end{table}
%
\par We examined how performance changed as we changed $P_{th}$ (see table \ref{table:param_consider} for description). Each edge in our flow network is associated with a probabilistic weight that represents how likely that edge transition is. Dropping edges whose weights are below $P_{th}$ from consideration is a method of handling outliers. The results are shown in table \ref{table:changing_k_p}. MTE values were typically the lowest for $P_{th} = .5$. With high $P_{th}$, low-quality edges get removed and the resulting trajectories are better probabilistically. 
\begin{table}
	\centering
	\caption{Outcome of changing $P_{th}$ on the normal data and MTE.}
	\begin{tabular}{|c|c|c|c|}
		\hline 
		$\mathbf{P_{th}}$ & \textbf{Overall/mm} & \textbf{ES/mm} & \textbf{ED/mm}  \\
		\hline  
		0.1 & 	0.92 $\pm$ 0.75 & 1.27 $\pm$ 0.79 &	0.83 $\pm$ 0.55 \\ \hline 
		0.3 & 	0.90 $\pm$ 0.70 & 1.24 $\pm$ 0.74 &	\textbf{0.80 $\pm$ 0.49} \\ \hline
		0.5	& 	\textbf{0.87 $\pm$ 0.63} & 1.23 $\pm$ 0.74 & 0.81 $\pm$ 0.51  \\ \hline
	\end{tabular}
	\label{table:changing_k_p}
\end{table}
%
\subsubsection{Effect of Different Constraints}
\begin{figure}
	\centering
	\subfloat[Overall MTE for all data.]{\includegraphics[scale=.30]{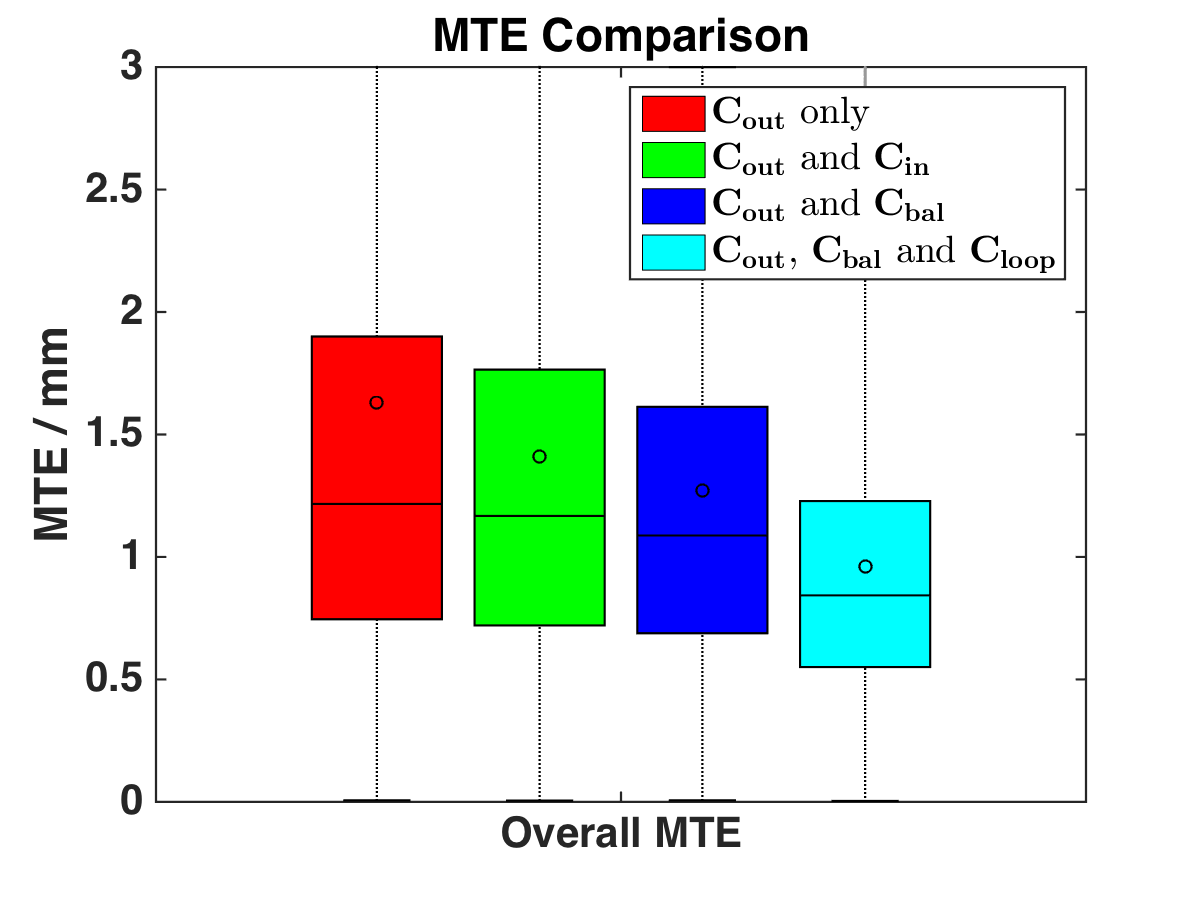}} \qquad
	\subfloat[MTE for ES and ED for all data.]{\includegraphics[scale=.30]{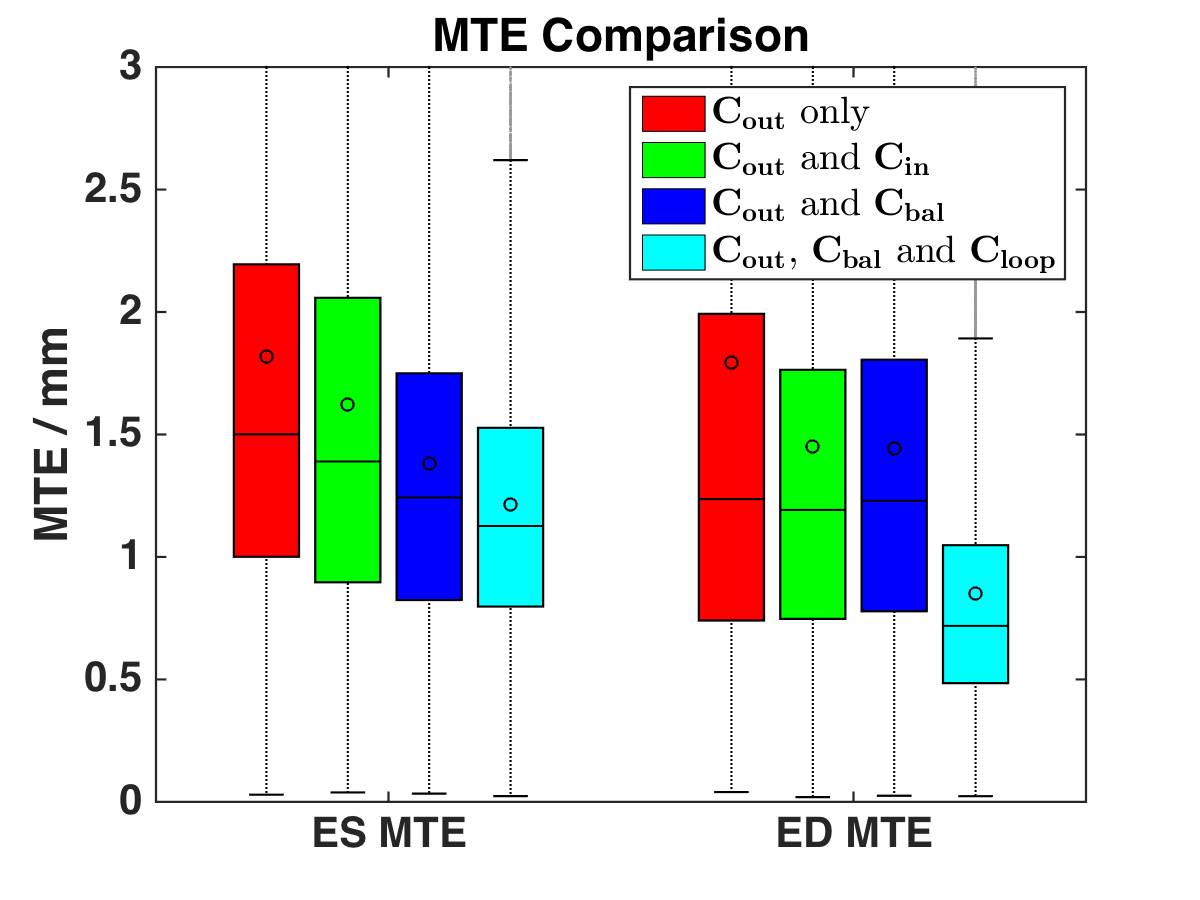}}
	\caption{MTE for all data, for different constraint setting. $C_{in}$, $C_{bal}$ and $C_{loop}$ were added incrementally.}
	\label{fig:mse_boxplot_w_wo_loop}
\end{figure}

Next, we see how the algorithm performs under different constraints. We start by applying $C_{out}$ only, then $C_{in}$, $C_{bal}$ and $C_{loop}$ in an incremental fashion. $C_{out}$ does not enforce any one-to-one correspondence constraint. It is equivalent to tracking all points independently using a shortest path formulation. $C_{out}$ and $C_{in}$ together enforce one-to-one correspondence at a frame-to-frame level. $C_{out}$ and $C_{bal}$ together enforce one-to-one correspondence throughout the cardiac cycle. Finally, applying $C_{loop}$ in addition to $C_{out}$ and $C_{bal}$ also enforces a balance between first and last frames, thereby encouraging trajectories to start and end at nearby positions. The findings are summarized in Table \ref{table:different_constraint_setting} and figure \ref{fig:mse_boxplot_w_wo_loop}. 
\par Not surprisingly, there is an incremental improvement with each additional constraint. The jump from having no loop constraint to the loop constraint is especially significant. This validates our intuition that accounting for the cyclical nature of cardiac motion is necessary. 
\begin{table}
	\centering
	\caption{Result of changing constraints applied to the optimization and MTE.}
	\begin{tabular}{|c|c|c|c|}
		\hline
		\textbf{Constraint configuration} & \textbf{Overall/mm} & \textbf{ES/mm} & \textbf{ED/mm} \\ \hline 
		$C_{out}$ only & 1.22 $\pm$ 1.15 & 1.50 $\pm$ 1.19 & 1.24 $\pm$ 1.25 \\ \hline
		$C_{out}$ and $C_{in}$ & 1.17 $\pm$ 1.04 & 1.39 $\pm$ 1.16 & 1.19 $\pm$ 1.02 \\ \hline
		$C_{out}$ and $C_{bal}$ & 1.09 $\pm$ 0.92 & 1.24 $\pm$ 0.93 & 1.23  $\pm$ 1.03 \\ \hline
		$C_{out}$, $C_{bal}$ and $C_{loop}$ & \textbf{0.84 $\pm$ 0.68} & \textbf{1.13 $\pm$ 0.73}  & \textbf{0.72 $\pm$ 0.56} \\ \hline	
	\end{tabular}
	\label{table:different_constraint_setting}
\end{table}
%
\subsubsection{Comparing Features}
\par
\begin{figure}
	\centering
	\subfloat[Overall MTE for all data.]{\includegraphics[scale=.32]{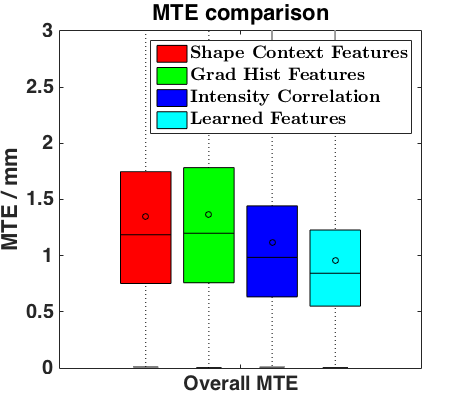}} \qquad
	\subfloat[MTE for ES and ED for all data.]{\includegraphics[scale=.32]{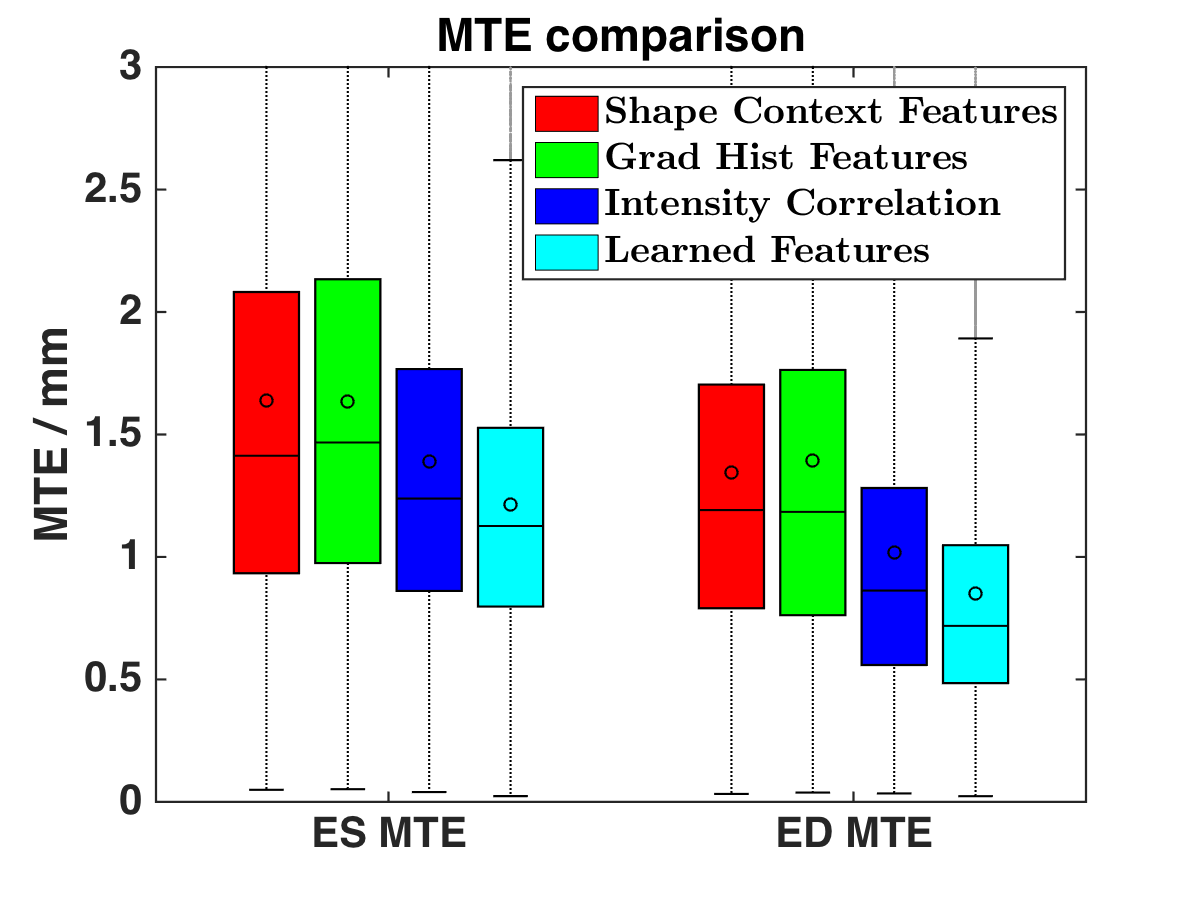}}
	\caption{MTE for all data, comparing different features. The same tracking method (FNT) was used for all of these.}
	\label{fig:mse_boxplot_w_sh_ctxt_and_lrnd_feats}
\end{figure}

 We also compared the performance of the learned (unsupervised) feature against other features (and/or metric). A comparison to the shape context feature \citep{belongie2000shape}, gradient histogram feature and intensity cross-correlation metric-based approaches is shown in table \ref{table:shpvslrnd} and figure \ref{fig:mse_boxplot_w_sh_ctxt_and_lrnd_feats}.
\begin{table}
	\centering
	\caption{Average MTE for different feature generation methods.}
	\begin{tabular}{|c|c|c|c|}
		\hline		
		\textbf{Feature} & \textbf{Overall/mm} & \textbf{ES/mm} & \textbf{ED/mm} \\ \hline 	
		Shape context & 1.19 $\pm$ 0.99 & 1.41 $\pm$ 1.15   & 1.19 $\pm$ 0.91 \\ \hline 
		Gradient histograms & 1.20 $\pm$ 1.03 & 1.47 $\pm$ 1.16  & 1.18 $\pm$ 1.00\\ \hline  
		Intensity Cross-correlation & 0.98 $\pm$ 0.81 & 1.24 $\pm$ 0.91 & 0.86 $\pm$ 0.72 \\ \hline    
		Learned feature (Autoencoder) & \textbf{0.84 $\pm$ 0.68} & \textbf{1.13 $\pm$ 0.73}  & \textbf{0.72 $\pm$ 0.56} \\ \hline  
	\end{tabular}
	\label{table:shpvslrnd}	
\end{table}
\par The learned feature using an Autoencoder provides better tracking results in comparison to other features generated using shape and image information. It was fascinating to find that the cross-correlation of intensity patches also performed relatively well. This is likely because speckle de-correlation is probably not substantial across time in this synthetic dataset. 
%

%

%
\subsubsection{Comparing Methods}
\par
\begin{figure}
	\centering
	\subfloat[Overall MTE for all data.]{\includegraphics[scale=.34]{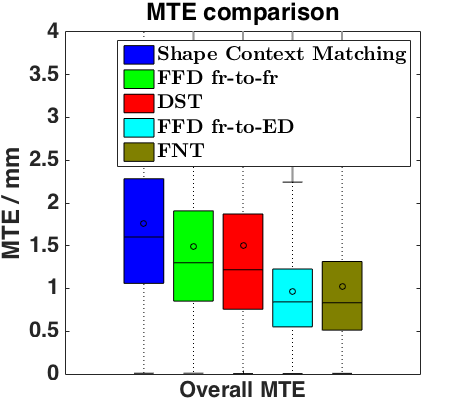}} \qquad
	\subfloat[MTE for ES and ED for all data.]{\includegraphics[scale=.34]{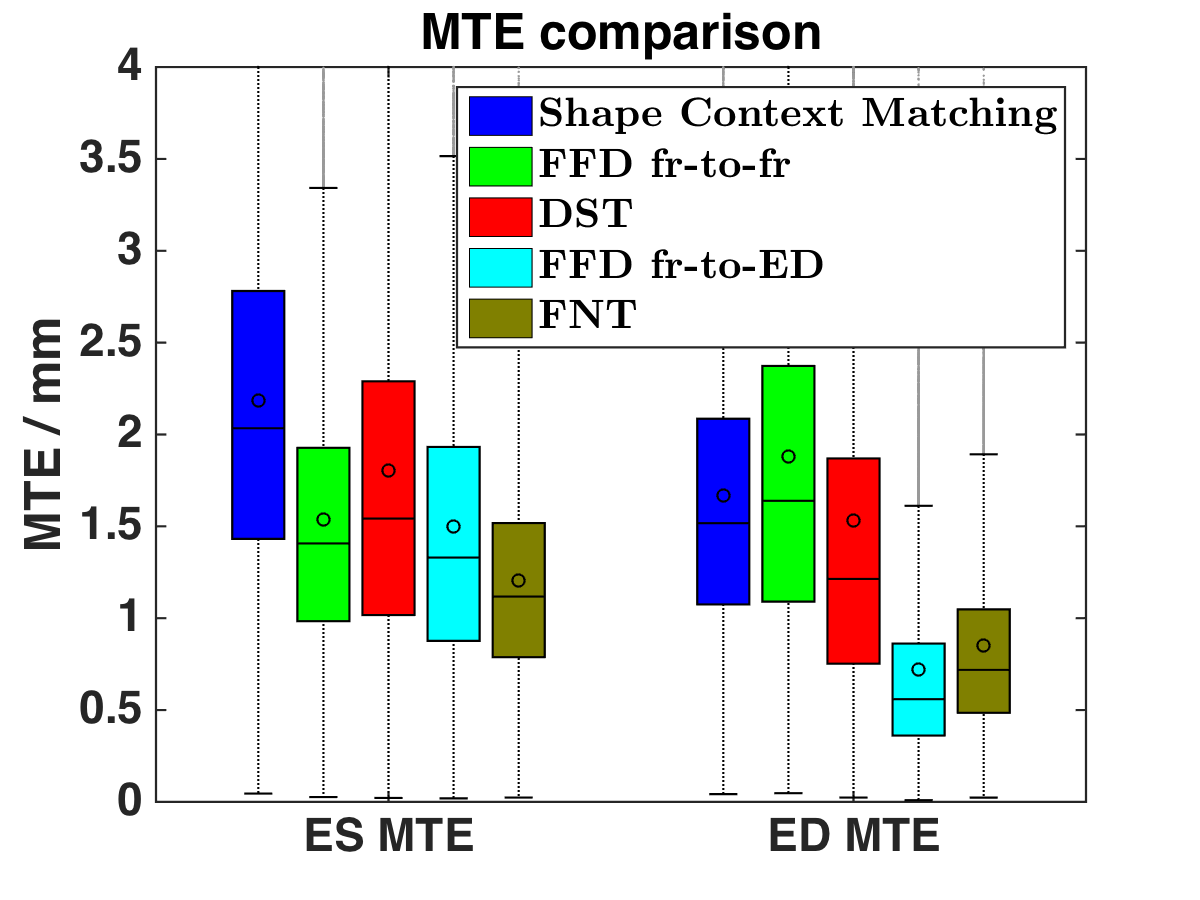}}
	\caption{MTE for all data, comparing different methods.}
	\label{fig:mse_boxplot_w_fnt_vs_others}
\end{figure}

 We compared the FNT method against other point tracking methods. Table \ref{table:fnt_vs_others} and figure \ref{fig:mse_boxplot_w_fnt_vs_others} summarize the findings. For shape context matching \citep{belongie2000shape}, we had to use a lower sampling rate ($Z_{fr} = 20$, $\theta_{fr} = 10$) because the algorithm took over an hour to run per frame and therefore was not tractable at a higher sampling rate. The Dynamic Shape Tracking (DST) algorithm is run with the same setting as the FNT algorithm ($Z_{fr} = 40$, $\theta_{fr} = 30$ and $NK=3$). A free-form deformation (FFD, \citet{rueckert1999nonrigid}) implementation available from the Bioimagesuite package \citep{bioimagesuite} was also applied to our data for reference. FFD was applied by registering all frames to the next frame in the sequence (FFD fr-to-fr) and also by registering all frames to the first frame (FFD fr-to-ED). 
\begin{table}
	\centering
	\caption{MTE for different tracking methods.}
	\begin{tabular}{|c|c|c|c|}
		\hline
		\textbf{Method} & \textbf{Overall/mm} & \textbf{ES/mm} & \textbf{ED/mm} \\ 
		\hline			
		Shape Context Matching &  1.60 $\pm$ 1.22 & 2.03 $\pm$ 1.35 & 1.52 $\pm$ 1.01 \\ \hline 
		DST & 1.22 $\pm$ 1.11 & 1.54 $\pm$ 1.27 & 1.21 $\pm$ 1.12  \\ \hline 
		FFD fr-to-fr & 1.30 $\pm$ 1.05 & 1.41 $\pm$ 0.94 & 1.64 $\pm$ 1.28   \\ \hline 
		FNT & 0.84 $\pm$ 0.68 & \textbf{1.13 $\pm$ 0.73}  & 0.72 $\pm$ 0.56  \\ \hline 
		FFD fr-to-ED & \textbf{0.83 $\pm$ 0.80} & 1.37 $\pm$ 1.08 & \textbf{0.56 $\pm$ 0.50}   \\ \hline 
	\end{tabular}
	\label{table:fnt_vs_others}	
\end{table}
\par We first note that the shape context matching and FFD applied frame-to-frame seem to provide the worst results overall. Since these methods did not have any temporal aspect to them, their tracking suffers a lot post systole, which is evident from the high ED errors. The DST method performs only slightly better as it does not track all points together but provides improved tracking post systole and therefore results in lower ED errors. The frame-to-ED FFD, where images are registered directly to the first frame and FNT seem to provide similar results overall. But FNT is better at ES. This is because the deformation from ED to ES is large and the FFD algorithm was most likely not able to account for that. Good performance at ES is crucial since peak strains typically occur around ES and are widely evaluated and reported clinically as a measure of function.
\subsubsection{Regional Strain Analysis}
\begin{figure}
	\centering
	\subfloat[Normal (FNT)]{\includegraphics[width=12cm, height=3.5cm]{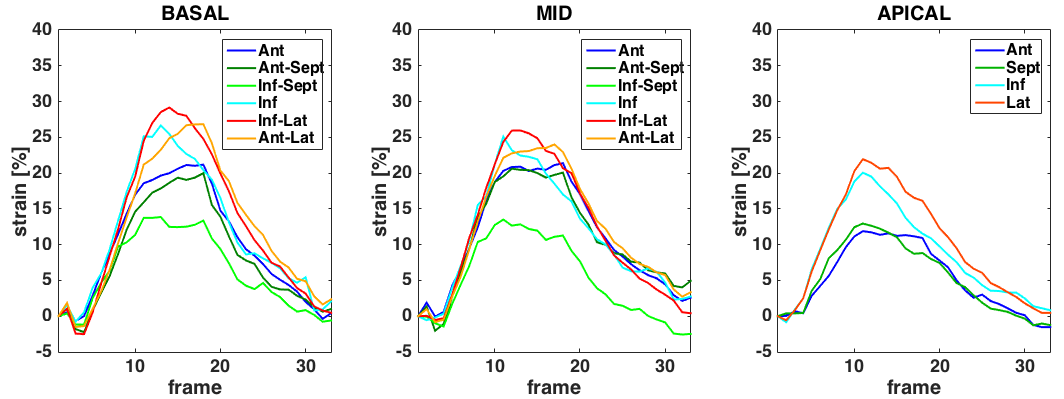}} \\
	\subfloat[Normal (Ground Truth)]{\includegraphics[width=12cm, height=3.5cm]{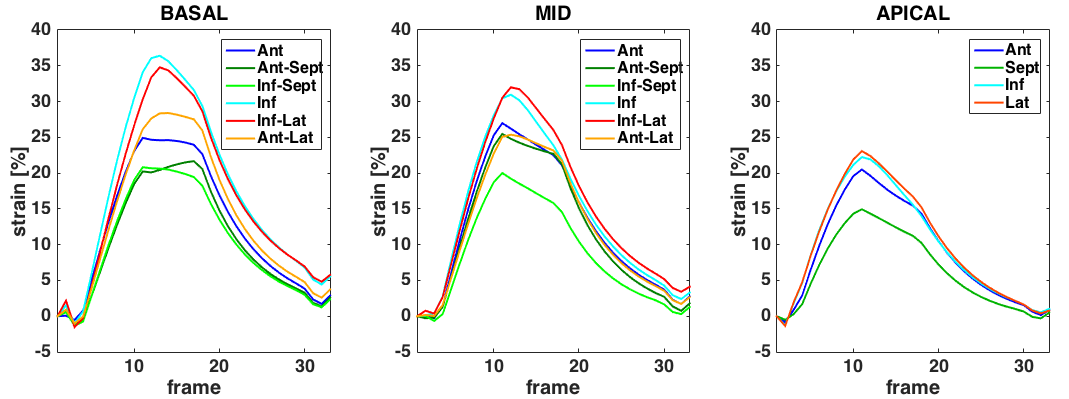}} \\
	\caption{Radial strain curves in the basal, mid and apical area of the LV for the normal Leuven data (our method and ground truth). Curves indicating mean strains for anterior (Ant) antero-septal (Ant-Sept), infero-septal (Inf-Sept), inferior (Inf), infero-lateral (Inf-Lat) and antero-lateral (Ant-Lat) regions are shown.}
	\label{fig:leuven_strain_normal_curves}
\end{figure}

Since we are ultimately interested detecting regional (within the LV) changes in strain values in order to isolate areas with injury, we also test whether the strain values we calculate can help us do this. Radial strain curves for the normal dataset, obtained using the FNT method and ground truth positions, are shown for basal, mid and apical areas of the LV in figure \ref{fig:leuven_strain_normal_curves}. Basal and mid areas are divided into 6 sectors - anterior (Ant), antero-septal (Ant-Sept), infero-septal (Inf-Sept), inferior (Inf),  infero-lateral (Inf-Lat) and antero-lateral (Ant-Lat) sectors. Apical area is divided into 4 sectors - anterior (Ant), septal (Sept), inferior (Inf) and lateral (Lat) sectors. For the normal data, we can see that strain values do not differ significantly across different sectors. Compared to the ground truth, our method seems to underestimate peak radial strain values as evident from figure \ref{fig:leuven_strain_normal_curves}. However, the broader trends seem to be consistent with the ground truth strain values. 
\par
\begin{figure}
	\centering
	\subfloat[LADPROX (FNT)]{\includegraphics[width=12cm, height=3.5cm]{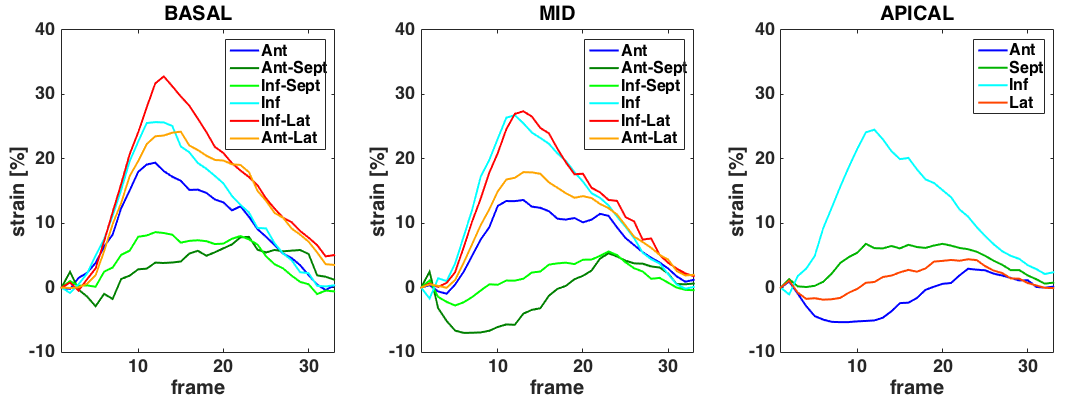}} \\
	\subfloat[LADPROX (Ground Truth)]{\includegraphics[width=12cm, height=3.5cm]{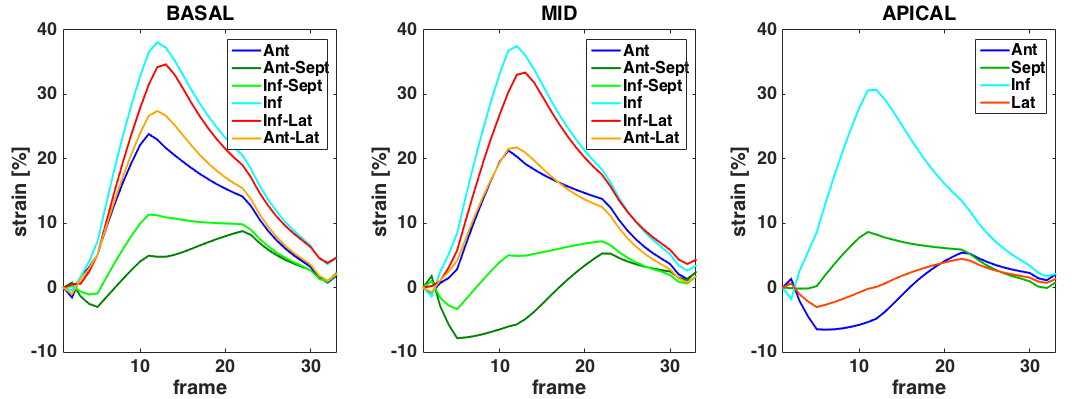}} \\
	\caption{Radial strain curves in the basal, mid and apical area of the LV for the LADPROX Leuven data (our method and ground truth). Curves indicating mean strains for anterior (Ant) antero-septal (Ant-Sept), infero-septal (Inf-Sept), inferior (Inf), infero-lateral (Inf-Lat) and antero-lateral (Ant-Lat) regions are shown.}
	\label{fig:leuven_strain_ladprox_curves}
\end{figure}

\begin{figure}
	\centering
	\subfloat[RCA (FNT)]{\includegraphics[width=12cm, height=3.5cm]{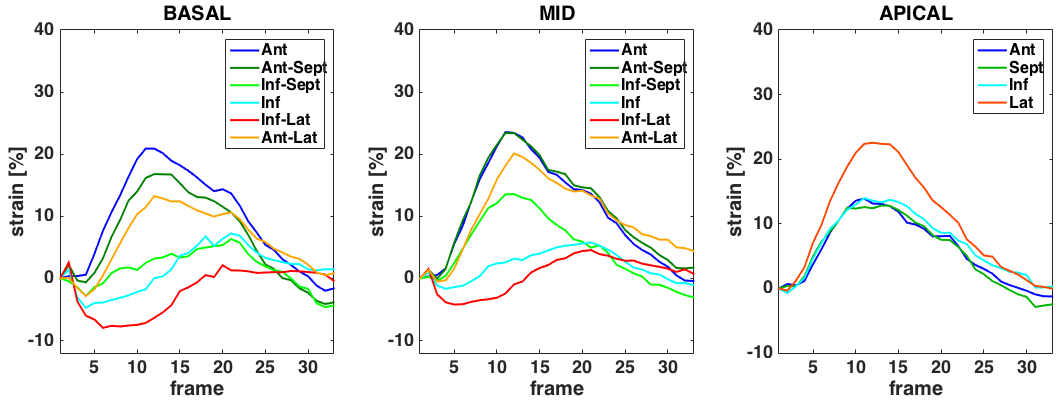}} \\
	\subfloat[RCA (Ground Truth)]{\includegraphics[width=12cm, height=3.5cm]{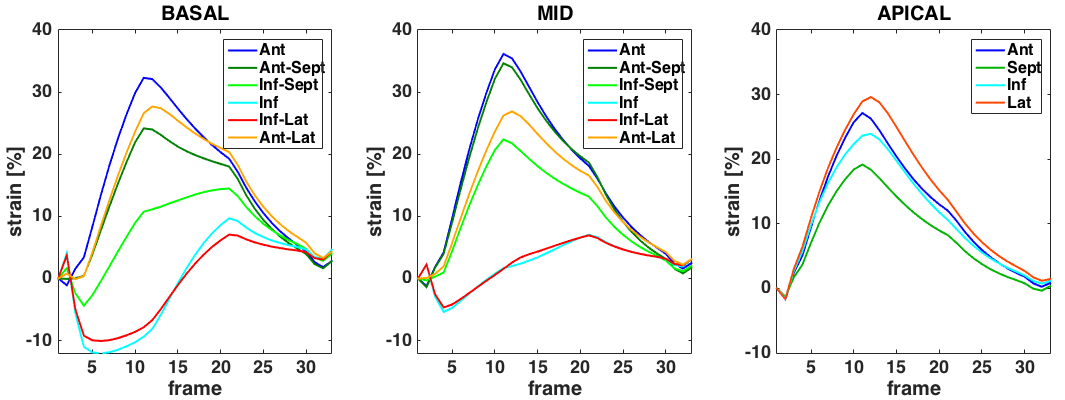}} \\
	\caption{Radial strain curves in the basal, mid and apical area of the LV for the RCA Leuven data (our method and ground truth). Curves indicating mean strains for anterior (Ant) antero-septal (Ant-Sept), infero-septal (Inf-Sept), inferior (Inf), infero-lateral (Inf-Lat) and antero-lateral (Ant-Lat) regions are shown.}
	\label{fig:leuven_strain_rca_curves}
\end{figure}

 In figure \ref{fig:leuven_strain_ladprox_curves}, radial strain curves for the LADPROX data are shown (FNT and ground truth position based). There appears to be a significant reduction of strain values around the infero-septal and infero-lateral sectors in the basal and mid regions. In the apical region, the anterior and lateral region strains are also lower. Figure \ref{fig:leuven_strain_rca_curves} shows radial strain curves for the RCA data (FNT and ground truth position based). Infero-lateral and inferior strain values are reduced in basal and mid regions. There appears to be no significant changes in the apical region. Similar to the normal data, there also appears to be an underestimation of strains for both the ischemic data. However, there is also a broad agreement on the strain trends. 
\par It should be clear from figures \ref{fig:leuven_strain_normal_curves}, \ref{fig:leuven_strain_ladprox_curves} and \ref{fig:leuven_strain_rca_curves} that, we are able to discern changes in strains across different regions of the LV. For instance, from normal to RCA, there is barely any change in the strain values in the apical area, whereas there is a significant reduction in inferior and infero-lateral strain values in both the basal and mid regions. From normal to LADPROX, there is a significant change in the apical strains. The observation that different injury profiles lead to different regional strain patterns, as demonstrated here, is of great value.
\par To quantify this more stringently, we directly compared strains obtained from FNT and the ground truth positions by analyzing the differences in strain values and cross-correlations. Table \ref{table:leuven_strain_abs_diff_by_group} summarizes absolute difference in radial, circumferential and longitudinal strain values for normal, Ischemic (LADPROX, LADDIST and RCA) and Dilated (LBBB, LBBBSMALL and SYNC) data groups. While the median absolute difference in strain values are within reasonable range for the normal and ischemic data, it is rather large for the dilated data group. This is perhaps indicative of the fact that we are not able to account for their complex motion patterns, which are different than that of the normal and ischemic group. 
\begin{table}
	\centering
	\caption{Median absolute difference of Lagrangian strains: Comparing FNT and ground truth.}
	\begin{tabular}{|c|c|c|c|}
		\hline 
		\textbf{Data group} & $\mathbf{Radial \quad (\%)}$ & $\mathbf{Circumferential \quad (\%)}$ & $\mathbf{Longitudinal \quad (\%)}$ \\ \hline
		\textbf{Normal} & 2.19 $\pm$ 2.94 & 1.37 $\pm$ 2.02 & 0.16 $\pm$ 0.21 \\ \hline 
		\textbf{Ischemic} & 2.39 $\pm$ 3.58 & 1.55 $\pm$ 2.24 & 0.18 $\pm$ 0.27 \\ \hline
		\textbf{Dilated} & 6.93 $\pm$ 9.69 & 3.79 $\pm$ 6.83 & 0.65 $\pm$ 1.12   \\ \hline  \hline
		\textbf{Overall} & 3.42 $\pm$ 5.85 & 2.12 $\pm$ 3.43 & 0.27 $\pm$ 0.53 \\   \hline 
	\end{tabular}
	\label{table:leuven_strain_abs_diff_by_group}
\end{table}
\par To understand whether these differences in strains were random or systematic, we also summarized median differences (not absolute) in table \ref{table:leuven_strain_diff_by_group}. This helps us identify the direction in which the algorithm is biased. The first thing of note is that, overall, there is no substantial bias in the circumferential and longitudinal strains. There is bias in all strain types for the dilated data group. Radial strains seem to be underestimated for all data groups, which is consistent with our findings from earlier as we observed the strain curves. 
%
\begin{table}
	\centering
	\caption{Median difference in Lagrangian strains: Comparing FNT and ground truth.}
	\begin{tabular}{|c|c|c|c|}
		\hline 
		\textbf{Data group} & $\mathbf{Radial \quad (\%)}$ & $\mathbf{Circumferential \quad (\%)}$ & $\mathbf{Longitudinal \quad (\%)}$ \\ \hline
		\textbf{Normal} & 2.13 $\pm$ 3.56 & 0.00 $\pm$ 2.75 & 0.06 $\pm$ 0.33 \\ \hline 
		\textbf{Ischemic} & 1.99 $\pm$ 4.19 & 0.00 $\pm$ 3.10 & 0.04 $\pm$ 0.36 \\ \hline
		\textbf{Dilated} & 5.18 $\pm$ 10.79 & -1.33 $\pm$ 7.69 & -0.03 $\pm$ 1.29   \\ \hline  \hline
		\textbf{Overall} & 2.79 $\pm$ 6.13 & -0.20 $\pm$ 4.54 & 0.02 $\pm$ 0.54 \\   \hline 
	\end{tabular}
	\label{table:leuven_strain_diff_by_group}
\end{table}
%
\par A reason behind this systematic bias in the radial strains is our use of segmented surfaces. Since points were constrained to move between surfaces, the maximum radial displacement was constrained. At a very small spatial scale, if we assume surfaces are flat (planes in 3D) and consecutive surfaces are parallel, the maximum possible radial motion is fixed - the projection of the normal vector between the two surfaces along the radial direction. However, because point correspondences are not perfect during optimization, there is noise in the displacement vectors. As these noisy vectors are regularized and smoothed, the final radial displacements are lower in aggregate than the ground truth. Such constraints do not exist circumferentially or longitudinally as long as points are sampled densely and regularly. 
\par Finally, we compare how well the trends agree with ground truth for the three strain types in table \ref{table:leuven_strain_correlations_by_group} by looking at the summary of correlation values of individual sector strain curves. Again, the dilated data seem to have the worst correlations, which is consistent with the findings so far.
Longitudinal strains are slightly worse and noisier overall. This is partly due to the fact that we use a sparse set of displacements located on the myocardial surfaces. Longitudinal motion is hard to quantify towards the basal and apical regions in this setting. Overall, the results with synthetic data were good both in terms of point tracking and strain analysis.
%
\begin{table}
	\centering
	\caption{Median correlations of Lagrangian strains: Comparing FNT and ground truth.}
	\begin{tabular}{|c|c|c|c|}
		\hline 
		\textbf{Data group} & $\mathbf{Radial}$ & $\mathbf{Circumferential}$ & $\mathbf{Longitudinal}$ \\ \hline
		\textbf{Normal} & 0.99 $\pm$ 0.02 & 0.96 $\pm$ 0.05 & 0.98 $\pm$ 0.04 \\ \hline
		\textbf{Ischemic} & 0.98 $\pm$ 0.04 & 0.96 $\pm$ 0.07 & 0.96 $\pm$ 0.15 \\ \hline
		\textbf{Dilated} & 0.87 $\pm$ 0.16 & 0.74 $\pm$ 0.35 & 0.42 $\pm$ 0.43 \\ \hline \hline
		\textbf{Overall} & 0.96 $\pm$ 0.10 & 0.93 $\pm$ 0.17 & 0.90 $\pm$ 0.56 \\   \hline 
	\end{tabular}
	\label{table:leuven_strain_correlations_by_group}
\end{table}

\par
\begin{figure}
	\centering
	\includegraphics[scale=.3]{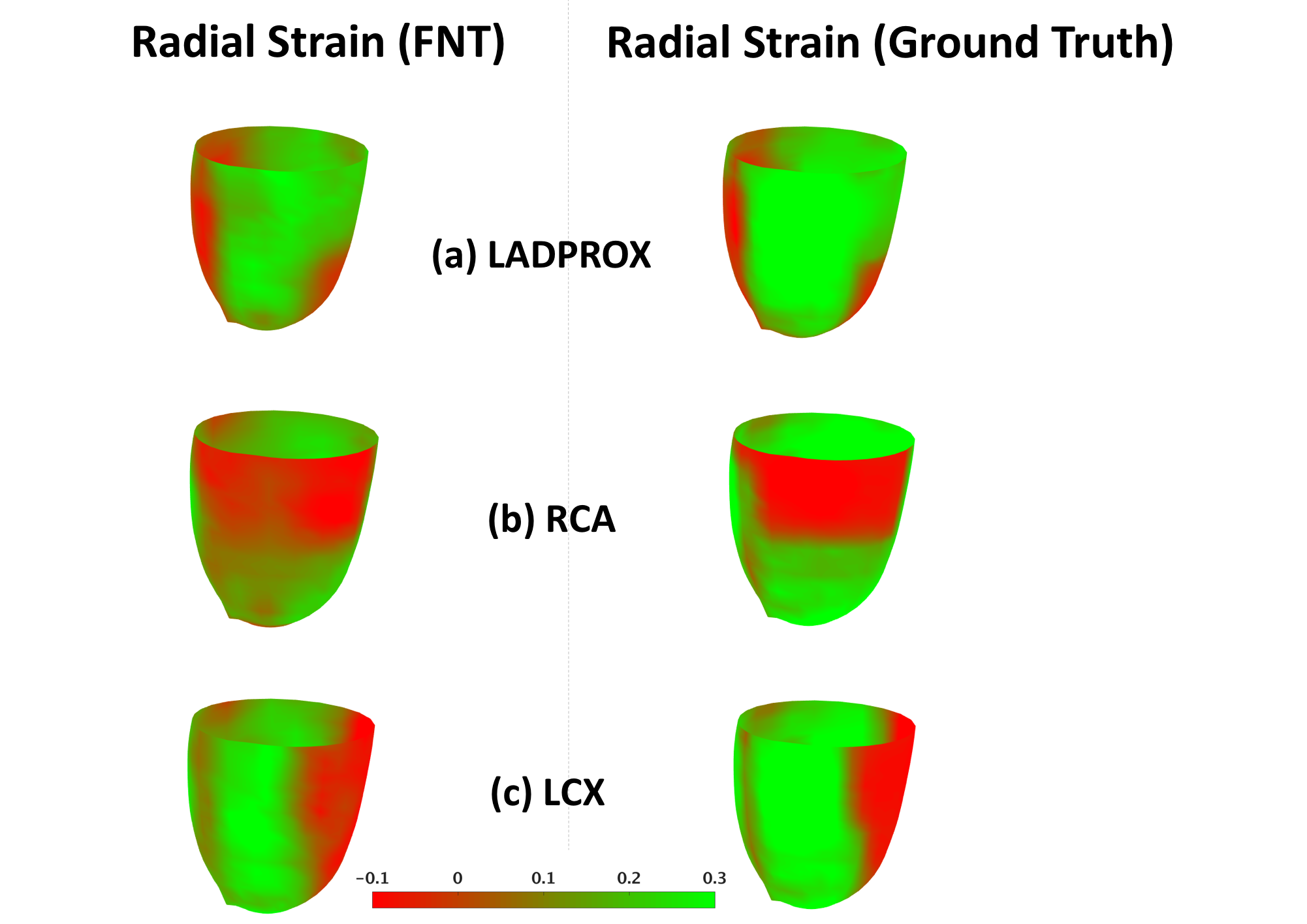}
	\caption{Epicardial surfaces displaying radial strains for three different types of ischemia, induced by occlusion at left anterior descending artery (LADPR0X), right coronary artery (RCA) and left circumflex artery (LCX).}
	\label{fig:strain_maps_example}
\end{figure}

 In figure \ref{fig:strain_maps_example} we also display radial strain maps at ES calculated using our method and the ground truth for 3 ischemic datasets. Areas with injuries have low strains and this can be seen in the strain maps. Again, this is to illustrate that we could reliably localize injuries since the maps from our method and ground truth appear fairly similar.
\subsection{\textit{In vivo} Data}
%
\subsubsection{Echocardiography Data}
We applied our method and explored physiological variations in the heart by analyzing \textit{in vivo} canine 4D (3D+t) echocardiography (4DE) data from 8 canine studies. The imaging was done on anesthetized open chested animals with a transducer suspended in a water bath. The animals were imaged in baseline condition (BL), an ischemic condition (high occlusion - HO), induced by occluding the left anterior descending artery (LAD),  and stress condition (HODOB), induced by infusing dobutamine at a low dosage: $5\mu g \backslash kg \backslash min$ in the presence of LAD ischemia. These conditions were tested due to our interest in ultimately developing a automated analysis of rest-stress images. Echocardiographic images were available for all 8 studies in all conditions but sonomicrometry data was available for 7 studies at BL and 5 studies during HO and HODOB.
%
\par Philips iE33 ultrasound system (Philips Medical Systems, Andover, MA), with the X7-2 phase array transducer and a hardware attachment that provided RF data, were used for acquisition. Imaging frequency ranged from 50-60 fps, which typically gave us 20-30 volumes per 4D image sequence. All experiments were conducted in compliance with the Institutional Animal Care and Use Committee policies.
\begin{table}
	\centering
	\caption{Different physiological conditions of imaging (\textit{in vivo} data).}
	\begin{tabular}{|p{2.7cm}|p{8cm}|}
		\hline 
		\textbf{Condition} & \textbf{Description} \\
		\hline
		BL &  Baseline. \\
		\hline 
		HO & High LAD occlusion. \\
		\hline 
		HODOB & High LAD occlusion with low dobutamine stress. \\
		\hline 		
	\end{tabular} 
	\label{table:different_conditions}
\end{table}
\begin{figure}
	\centering
	\includegraphics[scale=.29]{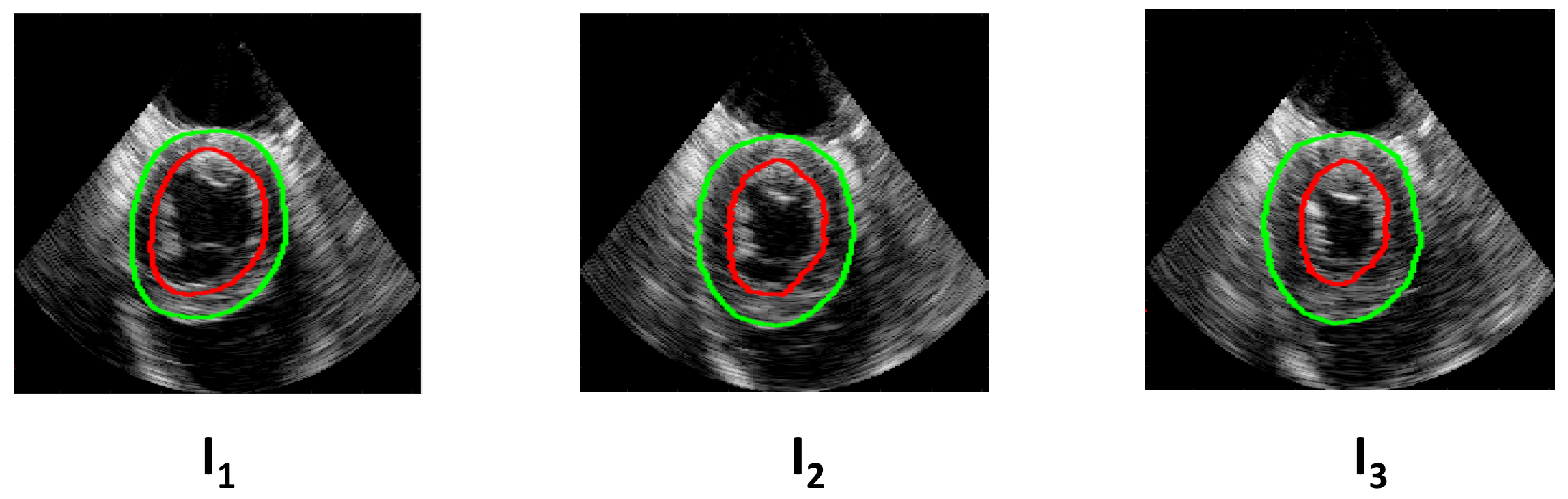}
	\caption{Example of \textit{in vivo} images and segmentation contours in one image sequence. $I_1$, $I_2$ and $I_3$ are three images in the systolic cycle.}
	\label{fig:in_vivo_example}
\end{figure}
\par Once these images were acquired, they were segmented using a semi-automated scheme. Endocardial and Epicardial surfaces were manually traced for the first frame of all images (see figure \ref{fig:in_vivo_example}). Then we used a dictionary learning based level set algorithm to propagate these surfaces through the cardiac cycle \citep{huang2014contour}. The FNT algorithm was then applied to these data with some adjustments. Since the extent of LV captured by imaging is different for different sequence in the long axis, $Z_{fr}$ is set as $ Z_{fr} = max(25, \text{total z slices available})$. $\theta_{fr}$ is then set as $\theta_{fr} = Z_{fr}/1.3$ since $1.3$ was the ratio of the best ($Z_{fr}$, $\theta_{fr}$) combination for the synthetic data - ($40, 30$). We used the unsupervised learning derived feature here as well, by training an autoencoder with \textit{in vivo} data. Other parameters were the same - $NK = 3$, $P_{th} = .5$. Radial basis function based interpolation method was used post FNT tracking and Lagrangian strains were calculated based on the techniques outlined in \citet{yan2007boundary}. 
\par We compared these strains with the ones obtained from sonomicrometric crystals implanted close to the mid-anterior LV wall during the same imaging studies as described above.
We focused on analyzing if the trends were consistent using correlation analysis. We describe the sonomicrometric crystal processing and calculations next. 
%
\subsubsection{Sonomicrometer Data}
We used sonomicrometric transducer crystals (crystals), recording instrument and processing software \textit{SonoSoft} and \textit{SonoView} (Sonometrics Corporation, London, Ontario, Canada) to acquire signal from crystals and process them. We implanted 19 crystals in the heart, 16 in the targeted areas of the left ventricle and 2 at the base in anterior and posterior locations and 1 at the true apex for a reference of the cardiac axis. Three additional crystals were placed in the edges of the transducer surface to align the crystals in echocardiographic LV coordinates. 
\par
\begin{figure}
	\centering
	\subfloat[Crystals arranged in 3 cuboidal lattices.]{\includegraphics[scale = .17]{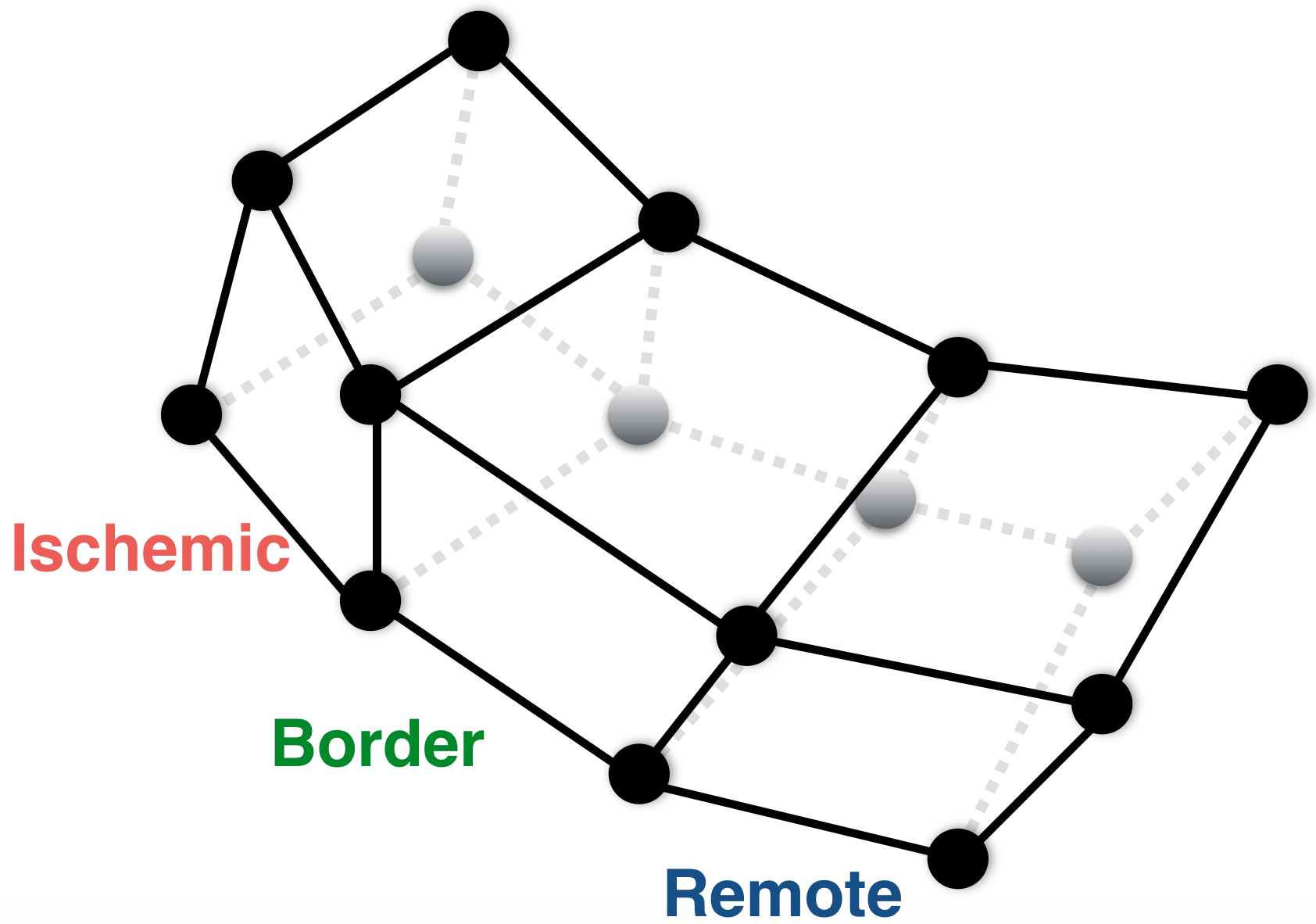}}
	\qquad \qquad \qquad 
	\subfloat[Crystal alignment in the LV.]{\includegraphics[scale = .35]{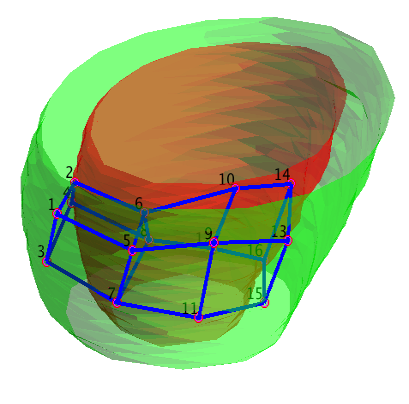}}
	\caption{Crystals and their relative position in the LV}
	\label{fig:crystal_placement}
\end{figure}

 The 16 crystals were arranged in such a way that they formed three adjacent cuboidal arrays. The three cubes were roughly located in: (i) the ischemic region of LV (ISC), which was caused by the aforementioned LAD occlusion (ii) the remote region (away from the ischemic area) and (iii) the border region between the two, as shown in figure \ref{fig:crystal_placement} (also table \ref{table:cube_names}).
\begin{table}
	\centering
	\caption{Cubical areas of crystal placement.}
	\begin{tabular}{|c|c|}
		\hline 
		\textbf{Area} & \textbf{Description} \\ \hline 
		\textbf{ISC} & Ischemic area. \\ \hline 
		\textbf{BOR} & Borderline area between ischemic and remote. \\ \hline 
		\textbf{REM} & Remote area. \\ \hline 
	\end{tabular}
	\label{table:cube_names}
\end{table}
%
\par We adapted the 2D sonomicrometry-based strain calculation method outlined in \citet{waldman1985transmural} for 3D. We calculated radial, circumferential and longitudinal strains using the apical and basal crystals that help define the cardiac geometry. We could then compare these strains with echocardiography (echo) based strains after aligning them in the LV coordinate system using rigid registration.
%
\subsubsection{Changes across Physiological Conditions}
\par
\begin{figure}
	\centering
	\subfloat[ISC]{\includegraphics[scale=.4]{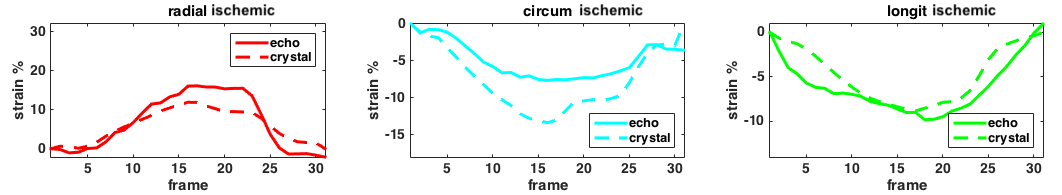}} \\
	\subfloat[BOR]{\includegraphics[scale=.4]{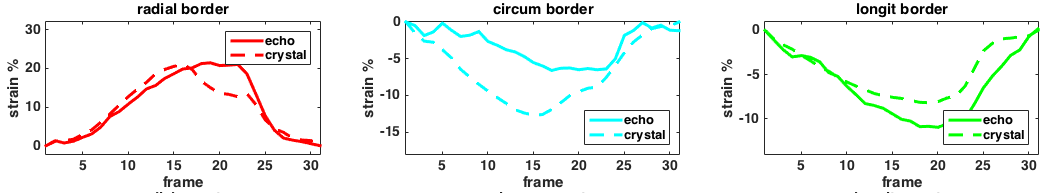}}	\\ 
	\subfloat[REM]{\includegraphics[scale=.4]{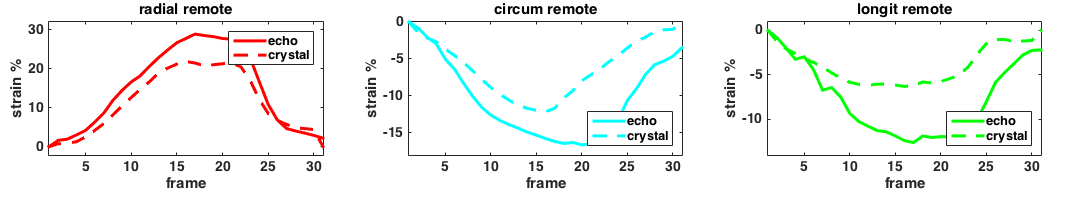}}
	\caption{FNT and crystal strains in BL condition for a data across the 3 cubic regions (ISC, BOR and REM top to bottom) for 1 dataset. Radial (red), circumferential (cyan) and longitudinal (green) strains from left to right.}
	\label{fig:dsea16_BL_comp_w_crystal}
\end{figure}

\begin{figure}
	\centering	
	\subfloat[ISC]{\includegraphics[scale=.4]{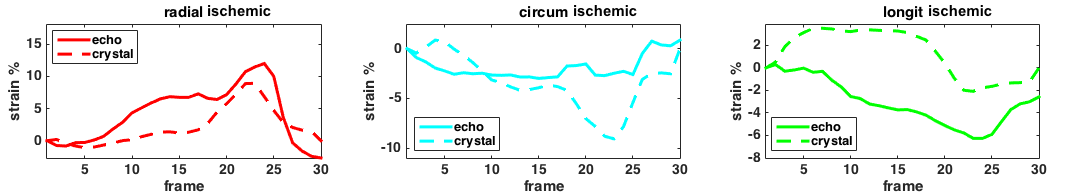}} \\
	\subfloat[BOR]{\includegraphics[scale=.4]{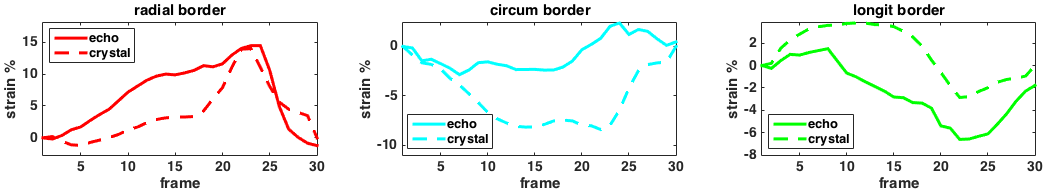}}	\\ 
	\subfloat[REM]{\includegraphics[scale=.4]{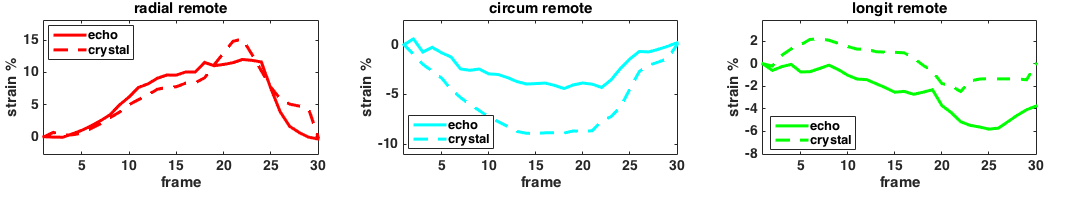}}
	\caption{FNT and crystal strains in HO condition for a data across the 3 cubic regions (ISC, BOR and REM top to bottom) for 1 dataset. Radial (red), circumferential (cyan) and longitudinal (green) strains from left to right.}
	\label{fig:dsea16_HO_comp_w_crystal}
\end{figure}

\begin{figure}
	\centering
	\subfloat[ISC]{\includegraphics[scale=.4]{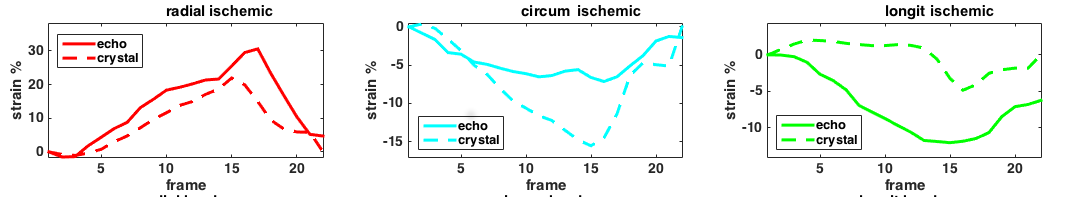}} \\
	\subfloat[BOR]{\includegraphics[scale=.4]{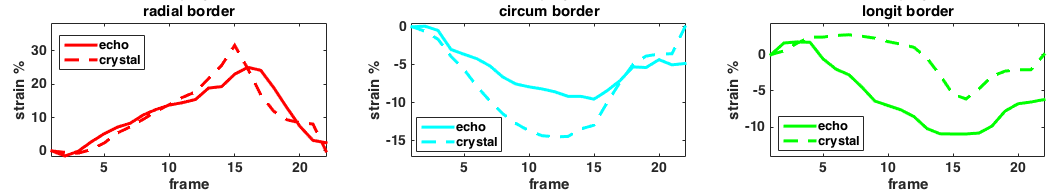}}	\\ 
	\subfloat[REM]{\includegraphics[scale=.4]{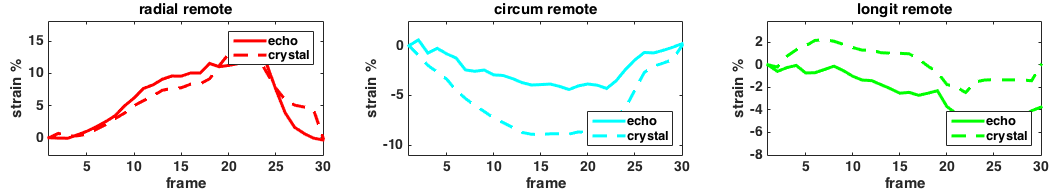}}
	\caption{FNT and crystal strains in HODOB condition for a data across the 3 cubic regions (ISC, BOR and REM top to bottom) for 1 dataset. Radial (red), circumferential (cyan) and longitudinal (green) strains from left to right.}
	\label{fig:dsea16_HODOB_comp_w_crystal}
\end{figure}

 We continued rest of our analysis by just using the FNT method. First, we explored how strains change from BL to HO and then to HODOB. We were primarily interested in exploring if the patterns of changes were different in the ischemic areas in comparison to the non-ischemic areas.
The ischemic (ISC) and non-ischemic areas were defined by the location of the crystals that defined the 3 cubic regions ISC, BOR and REM (see \ref{fig:crystal_placement}). Figures \ref{fig:dsea16_BL_comp_w_crystal}, \ref{fig:dsea16_HO_comp_w_crystal} and \ref{fig:dsea16_HODOB_comp_w_crystal} show strains for BL, HO and HODOB conditions respectively for 1 representative dataset.
\begin{table}
	\centering
	\caption{Median of peak \textbf{radial} strains for data across BL, HO and HODOB, also broken down by regions - ISC, BOR and REM. See figure \ref{fig:physio_response} for pictorial representation.}
	\begin{tabular}{|c|c|c|c|c|c|c|c|c|c|}
		\hline 
		\textbf{Method} & \multicolumn{3}{|c|}{\textbf{BL} $(\%)$} & \multicolumn{3}{|c|}{\textbf{HO}  $(\%)$} & \multicolumn{3}{|c|}{\textbf{HODOB}  $(\%)$} \\ \hline 
		\textbf{Crys} &  \multicolumn{3}{|c|}{12.3 $\pm$ 9.7} & \multicolumn{3}{|c|}{11.6 $\pm$ 7.1} & \multicolumn{3}{|c|}{30.2 $\pm$ 15.0} \\ \hline 
		\textbf{Echo} &  \multicolumn{3}{|c|}{13.7 $\pm$ 7.3} & \multicolumn{3}{|c|}{11.5 $\pm$ 4.4} & \multicolumn{3}{|c|}{22.0 $\pm$ 16.1} \\ \hline 
		& \textbf{ISC} & \textbf{BOR} & \textbf{REM} & \textbf{ISC} & \textbf{BOR} & \textbf{REM} & \textbf{ISC} & \textbf{BOR} & \textbf{REM}  \\ \hline 		
		\textbf{Crys} & 12.3 & 11.8 & 20.6 & 8.9 & 11.6 & 16.2 & 29.5 & 23.8 & 34.6  \\ \hline 		
		\textbf{Echo} & 16.0 & 15.7 & 13.0 & 12.0 & 11.3 & 12.0 & 22.0 & 25.0 & 19.5  \\ \hline 		
	\end{tabular}
	\label{table:physio_response_vals_rad}
\end{table}
\par There was a decrease in overall strain magnitudes, across all regions, going from BL (figure \ref{fig:dsea16_BL_comp_w_crystal}) to HO (figure \ref{fig:dsea16_HO_comp_w_crystal}). There was also recovery in all regions going from HO (figure \ref{fig:dsea16_HO_comp_w_crystal}) to HODOB (figure \ref{fig:dsea16_HO_comp_w_crystal}).
%
%
\par
\begin{figure}
	\centering
	\includegraphics[height=14cm, width=14cm]{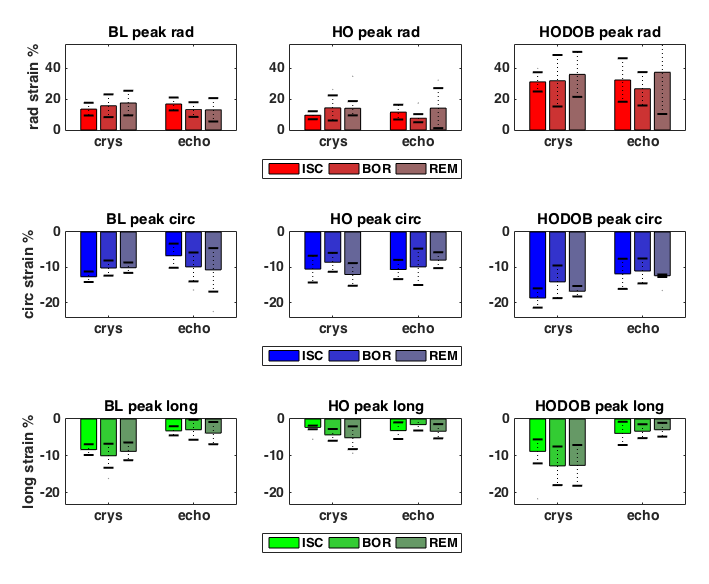}
	\caption{Peak strain bar graphs (with median and IQR) for radial (top), circumferential (middle) and longitudinal (bottom) strains at BL, HO and HODOB - shown across ISC, BOR and REM regions for echo and crystal-based strains.}
	\label{fig:physio_response}
\end{figure}

  Figure \ref{fig:physio_response} shows the changes in peak radial, circumferential and longitudinal strains from BL to HO to HODOB (for 8 BL data and 5 HO and HODOB data). Median values across different groups are shown, along with the IQR. The trend of decrease in peak strains from BL to HO and then the subsequent increase from HO to HODOB was strongly shown by radial strains. The corresponding radial strains can be found in table \ref{table:physio_response_vals_rad} as well. The change in HO to HODOB is more substantial for both crystal and FNT-based strains. There is a subtle difference in the magnitude of increase, from BL to HO, and decrease, from HO to HODOB, of radial strains across ISC, BOR and REM regions for the crystal-based strains. Such a difference was not found for the echo-based strains. 
\begin{table}
	\centering
	\caption{Median of peak \textbf{circumferential} strains for data across BL, HO and HODOB, also broken down by regions - ISC, BOR and REM. See figure \ref{fig:physio_response} for pictorial representation.}
	\begin{tabular}{|c|c|c|c|c|c|c|c|c|c|}
		\hline 
		\textbf{Method} & \multicolumn{3}{|c|}{\textbf{BL  $(\%)$}} & \multicolumn{3}{|c|}{\textbf{HO  $(\%)$}} & \multicolumn{3}{|c|}{\textbf{HODOB  $(\%)$}} \\ \hline 
		\textbf{Crys} &  \multicolumn{3}{|c|}{-11.2 $\pm$ 2.3} & \multicolumn{3}{|c|}{-10.2 $\pm$ 7.1} & \multicolumn{3}{|c|}{-16.4 $\pm$ 3.7} \\ \hline 
		\textbf{Echo} &  \multicolumn{3}{|c|}{-7.0 $\pm$ 2.8} & \multicolumn{3}{|c|}{-8.3 $\pm$ 3.4} & \multicolumn{3}{|c|}{-10.8 $\pm$ 3.1} \\ \hline 
		& \textbf{ISC} & \textbf{BOR} & \textbf{REM} & \textbf{ISC} & \textbf{BOR} & \textbf{REM} & \textbf{ISC} & \textbf{BOR} & \textbf{REM}  \\ \hline 		
		\textbf{Crys} & -12.8 & -10.0 & -10.7 & -12.1 & -12.1 & -10.4 & -19.3 & -14.5 & -16.4  \\ \hline 		
		\textbf{Echo} & -7.4 & -6.8 & -7.0 & -8.3 & -8.7 & -7.4 & -10.8 & -9.6 & -12.1  \\ \hline 		
	\end{tabular}
	\label{table:physio_response_vals_circ}
\end{table}
\par Circumferential strain values across the regions are reported in table \ref{table:physio_response_vals_circ}. Both crystal-based and echo-based (FNT) circumferential strains did not display changes in absolute magnitudes across ISC, BOR and REM (see figure \ref{fig:physio_response} and table \ref{table:physio_response_vals_circ}). Changes in longitudinal strains are also shown in figure \ref{fig:physio_response} and values reported in table \ref{table:physio_response_vals_long}. The crystal-based strains show decreases in strain magnitudes from BL to HO and increases from HO to HODOB. The echo strain magnitudes were very small and therefore not very informative since the IQRs were fairly high. Just like radial strains, the crystal-based longitudinal strains also point to the existence of a subtle difference in the magnitude of decrease from BL to HO, and increase from HO to HODOB, across ISC, BOR and REM regions. 
\begin{table}
	\centering
	\caption{Median of peak \textbf{longitudinal} strains for data across BL, HO and HODOB, also broken down by regions - ISC, BOR and REM. See figure \ref{fig:physio_response} for pictorial representation.}
	\begin{tabular}{|c|c|c|c|c|c|c|c|c|c|}
		\hline 
		\textbf{Method} & \multicolumn{3}{|c|}{\textbf{BL  $(\%)$}} & \multicolumn{3}{|c|}{\textbf{HO  $(\%)$}} & \multicolumn{3}{|c|}{\textbf{HODOB  $(\%)$}} \\ \hline 
		\textbf{Crys} &  \multicolumn{3}{|c|}{-9.1 $\pm$ 2.93} & \multicolumn{3}{|c|}{-3.2 $\pm$ 2.9} & \multicolumn{3}{|c|}{-11.6 $\pm$ 6.9} \\ \hline 
		\textbf{Echo} &  \multicolumn{3}{|c|}{-5.1 $\pm$ 3.2} & \multicolumn{3}{|c|}{-3.0 $\pm$ 5.2 } & \multicolumn{3}{|c|}{-4.2 $\pm$ 3.0} \\ \hline 
		& \textbf{ISC} & \textbf{REM} & \textbf{BOR} & \textbf{ISC} & \textbf{REM} & \textbf{BOR} & \textbf{ISC} & \textbf{REM} & \textbf{BOR}  \\ \hline 		
		\textbf{Crys} & -8.1 & -9.8 & -9.4 & -2.6 & -5.4 & -4.9 & -11.3 & -14.1 & -13.3  \\ \hline 		
		\textbf{Echo} & -5.1 & -4.9 & -5.7 & -3.0 & -2.6 & -5.8 & -3.9 & -4.7 & -3.3  \\ \hline 		
	\end{tabular}
	\label{table:physio_response_vals_long}
\end{table}
%
\par The results suggest that the overall pattern of changes in the echo-based strain magnitudes, aggregated over the three functional regions (ischemic, border and normal), are consistent with our expectations. In BL condition, we expected normal heart function and strain values. During HO, which induces ischemia, we expected an overall decrease in the heart function and strain magnitudes (primarily in the ischemic region). From HO to HODOB, we expected a recovery of function (primarily in the non-ischemic regions). However, the echo based strain results for the individual functional regions (ischemic, border and normal) were not distinctive enough. We expected there to be a greater decrease in strain magnitudes from BL to HO and smaller increase in strain magnitudes from HO to HODOB in ischemic areas (and vice versa for non-ischemic areas). While we were able to observe this to some extent with the crystal-based strains, that was not the case for the echo-based strains. 
\par A source of uncertainty in this crystal-based analysis was the challenge involved in registering the crystals with the LV. While the positions of the transducer were available from the reference crystals in the transducer, there still remained the task of rotational alignment that required some manual intervention. Furthermore, the transducer position crystals, which served as references, were themselves subject to noise and uncertainties. The ischemia that we induced was also possibly not severe enough to cause highly localized functional difference. Overall, even though the analysis of echo did not display highly localized sensitivity, the crystal strains did, which is a highly encouraging sign. 
\par 
\section{Conclusion}
In this work, we have developed, to the best of our knowledge, one of the first fully spatiotemporal cardiac surface tracking methods for 4DE data, which was then used to generate dense displacements using a regularized radial basis function framework. Our method was also able to account for the cyclical nature of the cardiac motion. Strains calculated from these displacements were then used to analyze the changes in global and local deformation of the LV.  We also proposed an unsupervised neural network-based feature generation method for motion tracking.
\par We obtained very good tracking accuracy with our method and features on the Leuven synthetic dataset. Even though we only tracked endocardial and epicardial surfaces, we obtained comparable results to the FFD method, which accounts for motion throughout the myocardium. We were also able to detect local regional changes in strain patterns and demonstrate expected strain profiles for different samples based on the location and extent of infarction.
%
%
\par In the future, the tracking method can be expanded to model dense myocardial displacements as well. This would have the advantage of not requiring perfect segmentations. A rough region-of-interest would suffice as long as it contains the LV. Since doing this would make the problem more ill-posed, the motion model would also have to be expanded to incorporate more constraints. The neural network-based feature generation strategy can be expanded further to develop supervised or transfer learning based methods applicable to \textit{in vivo} data as well. For instance, the Siamese neural network based approach we explored in our previous work can be extensively leveraged with more training data \citep{parajuli2017flow}.
\section{Acknowledgement} 
We are immensely thankful for the efforts of many past and present members of Dr. Albert Sinusas's group that were involved in the image acquisitions. We would also like to thank Dr. Hemant Tagare and Dr. Lawrence Staib of the Image Processing and Analysis Group at Yale for many fruitful discussions. This work was supported in part by the National Institute of Health (NIH) grant numbers R01HL121226 and T32HL098069.
\clearpage
\section*{References}
\bibliography{nripesh_manuscript}
\end{document}